\documentclass[lettersize,journal]{IEEEtran}
\usepackage{amsfonts}
\usepackage{amsmath}
\usepackage{algorithm}
\usepackage{algorithmicx}
\usepackage{algpseudocode}  % 这个和algorithmic不兼容，用了就要报错，好多莫名其妙的错误！！！！！
\usepackage{array}
\usepackage[caption=false,font=normalsize,labelfont=sf,textfont=sf]{subfig}
\usepackage{textcomp}
\usepackage{stfloats}
\usepackage{url}
\usepackage{verbatim}
\usepackage{graphicx}
\usepackage{cite}
\usepackage{hyperref} % 在导言区引入hyperref宏包

\usepackage{mdwmath}
\usepackage{mdwtab}
\usepackage{eqparbox}
\usepackage{tabularx}
\usepackage{adjustbox} % 调整表格
\usepackage{booktabs}
\usepackage{multirow}
\usepackage{diagbox}
\usepackage{float}
\usepackage{balance}

\usepackage{soul}
\usepackage{xcolor}
\definecolor{deepgreen}{rgb}{0.0, 0.8, 0.0} %
\usepackage{graphicx} % Required for inserting images
\usepackage{hyperref}
\hypersetup{
    urlcolor=blue,         % URL链接颜色
}

\title{Flying Co-Stereo: Enabling Long-Range Aerial Dense Mapping via Collaborative Stereo Vision of Dynamic-Baseline}

\author{Zhaoying~Wang${}^{1}$,
Xingxing~Zuo${}^{2,\star}$,
Wei~Dong${}^{1,\star}$% <-this % stops a space
\thanks{${}^{1}$ Zhaoying Wang and Wei Dong are with Shanghai Jiao Tong University, Shanghai 200240, China (e-mail: wangzhaoying@sjtu.edu.cn; dr.dongwei@sjtu.edu.cn;).
${}^{2}$ Xingxing Zuo is with the Department of Robotics, Mohamed Bin Zayed University of Artificial Intelligence (MBZUAI), Abu Dhabi, UAE. (xingxing.zuo@mbzuai.ac.ae)
${}^{\star}$ Corresponding authors with equal advising: Xingxing Zuo and Wei Dong.}
\thanks{}
}

\renewcommand{\thetable}{\Roman{table}}  % 改为大写罗马数字编号

\begin{document}

\maketitle

\begin{abstract} 
For Unmanned Aerial Vehicle (UAV) swarms operating in large-scale unknown environments, lightweight long-range mapping is crucial for enhancing safe navigation. Traditional stereo cameras constrained by a short fixed baseline suffer from limited perception ranges. To overcome this limitation, we present \textit{Flying Co-Stereo}, a cross-agent collaborative stereo vision system that leverages the wide-baseline spatial configuration of two UAVs for long-range dense mapping. However, realizing this capability presents several challenges. 
First, the independent motion of each UAV leads to a dynamic and continuously changing stereo baseline, making accurate and robust estimation difficult. Second, efficiently establishing feature correspondences across independently moving viewpoints is constrained by the limited computational capacity of onboard edge devices.
To tackle these challenges, we introduce the \textit{Flying Co-Stereo} system within a novel \textit{Collaborative Dynamic-Baseline Stereo Mapping (CDBSM)} framework. We first develop a dual-spectrum visual-inertial-ranging estimator to achieve robust and precise online estimation of the baseline between the two UAVs. In addition, we propose a hybrid feature association strategy that integrates cross-agent feature matching—based on a computationally intensive yet accurate deep neural network—with intra-agent, optical-flow-based lightweight feature tracking. 
Furthermore, benefiting from the wide baselines between the two UAVs, our system accurately recovers long-range co-visible 3D sparse points. We then employ a monocular depth network to predict up-to-scale dense depth maps, which are refined using accurate metric scales derived from the triangulated sparse points via exponential fitting.
Extensive real-world experiments demonstrate that the proposed \textit{Flying Co-Stereo} system achieves robust and accurate dynamic baseline estimation in complex environments while maintaining efficient feature matching with resource-constrained computers under varying viewpoints. Ultimately, our system achieves dense 3D mapping at distances of up to 70 meters with a relative error between 2.3\% and 9.7\%. This corresponds to up to a 350\% improvement in maximum perception range and up to a 450\% increase in coverage area compared to conventional stereo vision systems with fixed compact baselines.
The project webpage: \url{https://xingxingzuo.github.io/flying_co_stereo}
\end{abstract}

\begin{IEEEkeywords}
Flying Robots, Collaborative Visual Mapping, Dynamic Stereo Baseline, Real-Time Dense Mapping.
\end{IEEEkeywords}

\section{Introduction}

Onboard visual perception has become essential for autonomous Unmanned Aerial Vehicles (UAVs) to safely navigate unknown environments \cite{ren2025safety}, enabling applications from high-speed learning flight in the wild \cite{sciroboticsLearingFlight}, swarm flight in bamboo \cite{sciroboticsSwarmFlying}, and forest rescue operations \cite{tian2020search}. 
To ensure safe and fast navigation in large-scale environments, long-range perception capability is paramount \cite{ren2025safety}. Compared to LiDAR systems \cite{Zuo2019IROS, Zuo2020IROS, lv2021clins, nguyen2021viral, lang2023coco, lang2024gaussian}, cost-effective and lightweight stereo cameras can provide 3D depth perception through binocular disparity as well as richer visual information\cite{chen2021computationally}.
Conventional compact stereo cameras are typically limited to depth perception within 20 meters \cite{chen2021active}, constrained by their short fixed baselines. Although wide-baseline systems using multiple cameras on large fixed-wing aircraft have shown long-range mapping capabilities \cite{hinzmann2019flexible}, their platform size renders them unsuitable for small UAV swarms.

To overcome these limitations, we propose a dynamic wide-baseline collaborative stereo system that leverages visual cooperation between two UAVs. By forming a cross-agent stereo pair, the system achieves large inter-UAV parallax, enabling effective long-range dense mapping beyond the limits of traditional stereo setups.

Unlike conventional fixed-baseline systems \cite{sciroboticsLearingFlight} or static wide-baseline configurations \cite{gallup2008variableBaselineStereo}, our Flying Collaborative Stereo (Flying Co-Stereo) system distributes the stereo cameras across two dynamically flying UAVs. This introduces challenges such as dynamic, time-varying baselines and the difficulty of associating visual features across viewpoint-varying stereo image streams.

To address these challenges, we propose a \textit{Collaborative Dynamic-Baseline Stereo Mapping (CDBSM)} framework that integrates three tightly coupled components. 
First, we develop a \textit{Dual-Spectrum Visual-Inertial-Ranging Estimator (DS-VIRE)}, which fuses dual-spectrum visual observations (in both visible and infrared spectra) with inertial and UWB-based range measurements to achieve robust and precise online estimation of the dynamic baselines between two UAVs. 
Second, we design a hybrid feature association strategy that integrates cross-agent deep feature matching with intra-agent optical-flow-based feature tracking, ensuring real-time and persistent co-visible feature association across cameras under changing viewpoints.
Finally, the estimated dynamic baselines, along with the associated co-visible features, are leveraged to accurately recover metric 3D sparse landmarks—even at long ranges. These 3D sparse landmarks are further used to enrich the up-to-scale monocular dense depth predictions from a foundational model with accurate metric scaling, ultimately enabling precise metric dense mapping at long distances.

The main contributions of this article can be summarized as follows:
\begin{itemize}
  \item A Flying Co-Stereo system is proposed, in which two collaborative UAVs form a wide-baseline, cross-agent stereo vision setup within a unified Collaborative Dynamic-Baseline Stereo Mapping (CDBSM) framework, enabling long-range dense mapping in large-scale unknown environments.
  \item A dual-spectrum visual-inertial-ranging estimator is developed to achieve robust and accurate online estimation of the dynamic inter-UAV baseline in complex outdoor conditions.
  \item A hybrid visual feature association strategy is designed, combining cross-agent deep matching with intra-agent feature tracking, to ensure real-time and persistent co-visible feature correspondences under varying viewpoints.
  \item A sparse-to-dense depth recovery scheme is proposed, which refines dense monocular depth predictions using exponential fitting of long-range triangulated sparse landmarks for precise metric-scale mapping.
\end{itemize}

\section{Related Work}
\subsection{Wide-Baseline Stereo Vision}
The evolution of wide-baseline stereo vision systems can be traced through several key developments. The foundational work by Gallup et al.\cite{gallup2008variableBaselineStereo} first demonstrate that extending baseline lengths beyond traditional stereo configurations could significantly enhance long-range mapping resolution. Building on this concept, Hinzmann et al.\cite{hinzmann2019flexible} propose a practical wide-baseline system by mounting three cameras on a fixed-wing aircraft for long-range mapping during forward-flight. Achtelik et al.\cite{achtelik2011collaborative_stereo} advances the field by simulating a collaborative stereo using two UAVs, theoretically investigating baseline estimation through common visual features and IMU data. Karrer et al.\cite{karrer2021distributed} further develop this approach through simulating a wide-baseline downward stereo for high-altitude Simultaneous Localization and Mapping (SLAM). 
While these works have significantly advanced wide-baseline stereo vision, they remain predominantly theoretical and simulation-focused, with limited experimental validation in real-world scenarios. Wang et al.\cite{wang2024collaborative} first implement a practical wide-baseline stereo vision system using two UAVs. However, achieving accurate long-range dense mapping with wide baselines on collaborative flying UAVs remains an open challenge in this domain.

\subsection{Online Estimation of Dynamic Stereo Baseline}
Since the stereo cameras are mounted on two independently flying UAVs operating outdoors, robust and accurate online estimation of extrinsic parameters between left and right cameras becomes essential. Given the rigid camera-to-UAV body attachment, this estimation problem can be effectively transformed into relative pose estimation between the UAVs. 
 
Early studies\cite{zhang_agile_2022,Kalaitzakis2021} implement direct relative pose estimation through mutual observation of visual fiducial markers (ArUco, ARTag, AprilTag) mounted on UAVs. Subsequent developments\cite{VisionDroneFlocking,nguyen2020vision} advance this approach through CNN-based detection of neighboring UAVs for bearing observation. Building upon this foundation, recent advancements\cite{pavliv_headset_tracking_2021} improve pose estimation accuracy by incorporating semantic keypoint detection\cite{pavlakos_6dof_semantic_keypoints} following the initial CNN-based recognition. However, these visual approaches in the visible spectrum remain inherently vulnerable to outdoor illumination variations and complex backgrounds. To improve outdoor robustness, researchers have developed infrared-based solutions. The fundamental approaches\cite{VI-RPE-2018,wang_robust_2022} utilize infrared LED markers mounted on UAVs for relative position estimation through Perspective-n-Point (PnP) solutions. Parallel to infrared developments, alternative approaches\cite{walter_uvdar_2019} utilizing ultraviolet (UV) markers have been proposed to improve background differentiation in challenging lighting conditions.

CREPES\cite{xun2023crepes} further introduces the fusion of distance measurement from Ultra-Wideband (UWB) and infrared markers to enhance the robustness of relative pose estimation. Expanding the sensor suite further, Xu et al.\cite{xu2020decentralized} develop a comprehensive framework integrating individual odometry, UWB measurements, mutual observations, and common environmental features for high-precision estimation. Subsequent advancements lead to Omni-Swarm\cite{xu2022omni} with fisheye cameras and D2SLAM\cite{xu2024d2slam} with quad-fisheye configurations for inter-UAV relative pose estimation. While these systems achieve remarkable accuracy in indoor environments, their exclusive reliance on cameras of the visible spectrum may limit robustness in outdoor scenarios with potential light noises and complex backgrounds.

\subsection{Cross-Camera Feature Association}
Stereo camera systems require efficient and precise feature correspondence establishment between binocular views. Traditional methods such as SGM\cite{hirschmuller2007stereo} and SGBM\cite{geiger2010efficient} depend on highly accurate extrinsic parameters, making them suitable for fixed-baseline configurations. While sparse descriptor-based feature association proves to be a more efficient and viable approach for the dynamic varying stereo configuration\cite{zou2012coslam}.
Early multi-robot SLAM systems, such as CVI-SLAM\cite{karrer2018cvi}, CCM-SLAM\cite{schmuck2019ccm}, COVINS\cite{schmuck2021covins}, and CoSLAM\cite{zou2012coslam}, primarily employ ORB detectors\cite{murartal_orb-slam2_2017} for cross-camera feature matching using nearest neighbor (NN)\cite{david2004SIFT} strategies. While SURF\cite{bay2006surf} offers improved matching accuracy, its computational complexity results in limited real-time performance. 
Subsequent advances have shown that learned detectors such as SuperPoint\cite{detone2018superpoint} and learned matchers like SuperGlue \cite{sarlin2020superglue}, LightGlue\cite{lindenberger2023lightglue}, and LoFTR\cite{sun2021loftr} provide superior robustness to large viewpoint changes compared to classical methods. Omni-Swarm \cite{xu2022omni} and D2SLAM \cite{xu2024d2slam} employ an attention-based SuperPoint detector \cite{detone2018superpoint} for inter-UAV co-visible feature matching. However, this matching process is time-consuming on resource-constrained onboard systems and thus can only be performed intermittently rather than continuously in real-time. Although Zhu et al.\cite{zhu2021distributed} demonstrate the benefits of persistent multi-frame co-visible feature utilization for collaborative SLAM in simulation studies, achieving continuous and real-time inter-UAV feature association in practice remains an open challenge.

\subsection{Long-Range Dense Mapping}
SLAM techniques serve as early approaches for dense 3D structure reconstruction. Notable methods such as SVO\cite{forster2014svo}, REMODE\cite{pizzoli2014remode}, and LSD-SLAM\cite{engel_LSD_SLAM} achieve dense or semi-dense mapping via photometric alignment of motion-stereo image sequences. However, long-range depth estimation requires sufficient parallax, which undermines photometric consistency under large viewpoint changes. For navigation tasks that require dense geometric maps, Kimera\cite{rosinol2020kimera} and Teixeira et al.\cite{teixeira2016real} adopt simpler sparse-feature-based Visual-Inertial Odometry (VIO) pipelines, generating dense reconstructions via Delaunay triangulation of sparse landmarks. To further enhance mapping resolution, CNN-SLAM\cite{tateno2017cnnslam} incorporates single-view depth prediction from CNN networks for dense map reconstruction. However, the high-dimensional optimization of dense depth is computationally intensive. To reduce complexity, CodeSLAM\cite{bloesch2018codeslam}, CodeMapping\cite{matsuki2021codemapping}, and CodeVIO\cite{zuo2021codevio} employ a CVAE network architecture to encode a latent code for dense depth prediction from a single RGB image. However, the low-dimensional latent code is highly dataset-specific and lacks generalizability to other datasets. In addition, these approaches are only effective with near-field and abundant feature landmarks. The performance significantly degrades in large-scale, long-range depth environments.

Multi-view stereo (MVS) neural networks aim to reconstruct dense depth from posed reference and neighboring images. Approaches like SimpleRecon\cite{sayed2022simplerecon}, and SimpleMapping\cite{xin2023simplemapping} typically construct plane-sweep cost volumes from multiple frames, followed by 3D or 2D regularization. However, existing MVS methods largely depend on annotated datasets such as ScanNet\cite{dai2017scannet} for training, and their experimental validation has been primarily limited to indoor or small-scale outdoor environments. Recently, zero-shot approaches like MVSAnywhere\cite{izquierdo2025mvsanywhere} have demonstrated the potential of generalizing MVS reconstruction to arbitrary scenes and depth ranges. Nevertheless, such methods exhibit a high sensitivity to the accuracy of input image poses.

Sparse-to-dense depth completion is also a promising approach, which utilizes the RGB image and the sparse 3D point samples (from lidar or VIO landmarks) to predict dense depth maps. The research methods encompass traditional geometry\cite{ma2016sparse}, as well as self-supervised or unsupervised neural networks based on photometric consistency, temporal sequences of adjacent frames, and geometric constraints\cite{wong2020unsupervised,teixeira2020aerial}. These methods are typically trained on datasets like VOID\cite{wong2020VOID}, NYUv2\cite{NYUv2_2012}, and KITTI\cite{Geiger2013KITTIDataset}, where they perform well, but their generalization ability is also limited in large-scale and diverse outdoor scenes. Adapting them to large-scale outdoor environments requires retraining, yet such datasets are scarce.

Leveraging large pre-trained monocular depth models has emerged as a feasible approach, with Transformer-based models like MiDaS\cite{ranftl2020MiDas} and DepthAnythingV2\cite{yang2024DepthAnythingV2} demonstrating strong multi-scene generalization in monocular depth estimation.
However, monocular depth estimation inherently lacks precise scale information. NeRF-VO\cite{naumann2024nerfvo} addresses this limitation by linearly fitting sparse feature points to the DPT\cite{ranftl2021ViT} monocular depth estimates and optimizing the 3D structure within a Neural Radiance Fields (NeRF) framework. While this achieves high reconstruction accuracy, the real-time capability of NeRF is limited, making it unsuitable for onboard real-time applications. To improve efficiency, AB-VINS\cite{merrill2024ABVINS} uses VIO sparse feature points to linearly fit the scale and bias of monocular depth estimates from MiDaS\cite{ranftl2020MiDas}, constructing dense depth maps. This approach performs well in small-scale indoor scenes. However, in large-scale outdoor environments with significant depth variations between foreground and background regions, simple linear fitting struggles to achieve accurate reconstruction across both near and far fields, warranting further research.

\section{SYSTEM OVERVIEW}

\subsection{Terminology and Abbreviations}

\begin{itemize}
  \item \textbf{Flying Co-Stereo}: A collaborative stereo vision system consisting of two cameras mounted on two flying UAVs.
  
  \item \textbf{CDBSM}: A Collaborative Dynamic-Baseline Stereo Mapping framework, systematically designed for the Flying Co-Stereo to enable long-range dense mapping.

  \item \textbf{DS-VIRE}: A Dual-Spectrum Visual-Inertial-Ranging Estimator, as a core component of the CDBSM framework, supporting robust and precise online baseline estimation.

  % \item \textbf{DS-MVDT}: A Dual-Spectrum Marker-based Visual Detection and Tracking algorithm, embedded in DS-VIRE as a submodule of CDBSM, responsible for robust online baseline \textit{position} estimation in Flying Co-Stereo.

  \item \textbf{BVD}: A Bidirectional Visual Differential algorithm, embedded in DS-VIRE as part of the CDBSM framework, especially supporting accurate orientation estimation of the baseline.

  \item \textbf{GP-SS}: A Guidance-Prediction SuperPoint-SuperGlue algorithm, serving as the cross-agent co-visible feature association module within the CDBSM framework, responsible for efficient binocular feature matching in Flying Co-Stereo under dynamic viewpoints.
\end{itemize}

\subsection{Symbol Definitions}
\begin{description}
  \item[$(\cdot)^k$] Denotes the $k$-th frame captured by the front camera of UAV0, with $t_k$ as its timestamp. The system uses this camera’s frame time as the temporal reference.
  \item[$\mathbf{p}_{01}^{k}$] Relative position (translation) from UAV0 to UAV1 at $t_k$ timestamp.
  \item[$\mathbf{R}_{01}^{k}$] Relative rotation from UAV0 to UAV1 at $t_k$ timestamp, which is also represented by euler angles as $\mathbf{R}_{01} = [\phi_{01}, \theta_{01}, \psi_{01}]^\mathrm{T}$.
  \item[$C_0^k$] Front camera frame of UAV0 at $t_k$ timestamp.
  \item[$C_1^k$] Front camera frame of UAV1 at $t_k$ timestamp.
  \item[$f_m^k$] 2D features observed in the front camera frame $C_m^k, m \in(0,1)$.
  \item[${}^{C_m^k} \mathbf{p}_f $] 3D position of visual landmark (feature) in $C_m^k$ frame coordinates.
  \item[${}^A \mathbf{p}_f $] 3D position of  visual landmark (feature)  in anchor ${}^A(\cdot)$ frame coordinates.
  \item[$(u, v)$] 2D image observation of co-visible landmark (pixel coordinates)
  \item[$l$] Baseline length between two UAVs.
\end{description}

\subsection{Hardware Setup}
\begin{figure*}[]
  \centering
  \includegraphics[width=1.0\linewidth]{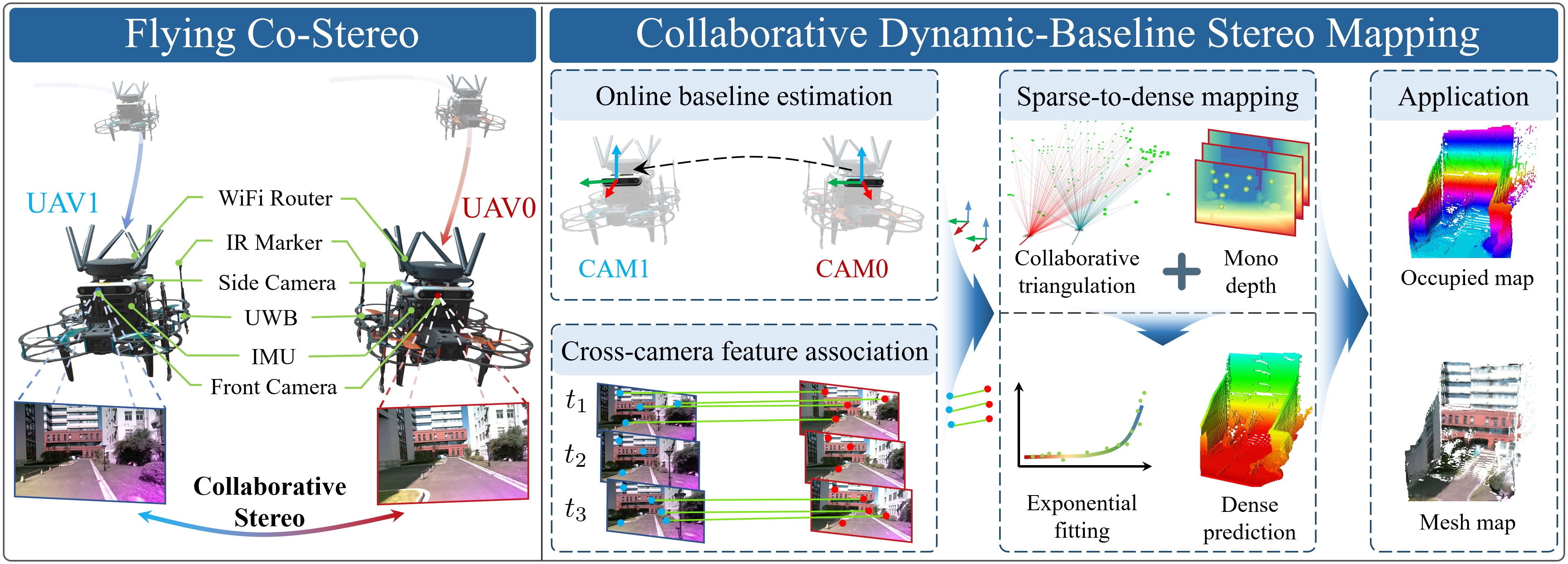}
  \caption{The system architecture of Flying Co-Stereo within our proposed Collaborative Dynamic-Baseline Stereo Mapping framework. The online baseline estimation module estimates the relative pose between flying stereo cameras. The cross-camera feature association module delivers co-visible feature correspondences. Building upon the poses and features, the sparse-to-dense mapping module first triangulates co-visible features into sparse landmarks, and then predicts dense mapping by exponentially fitting with monocular depth estimation. The multi-frame dense prediction map can be further processed through TSDF fusion to generate mesh maps and occupied maps.}
  \label{fig:1_system}
\end{figure*}

The Flying Co-Stereo system consists of two UAVs flying in parallel, as shown in the left part of Fig.~\ref{fig:1_system}. UAV0 acts as the leader, while UAV1 follows. Each UAV is equipped with a forward-facing camera, together forming the left and right views of a dynamic, wide-baseline stereo vision system. To estimate the time-varying baseline, side-mounted cameras and IR LED markers enable mutual visual observations between the UAVs. UWB modules installed on each UAV provide relative distance measurements. IMU data is utilized for both self Visual-Inertial Odometry (VIO) and relative baseline estimation. High-bandwidth, low-latency communication is ensured by meshed WiFi routers mounted atop each UAV. 

\subsection{Software Framework}
The Collaborative Dynamic-Baseline Stereo Mapping (CDBSM) framework is illustrated on the right side of Fig.~\ref{fig:1_system}. The online baseline estimation module provides robust and accurate estimation of stereo camera poses. The cross-camera feature association module adopts a guidance-prediction strategy to deliver real-time and persistent co-visible feature correspondences. Based on the wide-baseline stereo poses and co-visible feature observations, the sparse-to-dense mapping module first triangulates long-range sparse landmarks and then enriches up-to-scale monocular depth with metric sparse landmarks via exponential fitting to obtain the dense depth map.
For downstream applications, the multi-frame dense prediction map can be further processed through TSDF fusion\cite{oleynikova2017voxblox}, generating mesh maps and occupied maps.

\section{Flying Co-Stereo Baseline Estimation}
Relative pose estimation between the two cameras on two UAVs is a core component of the CDBSM framework. To establish the poses of the two cameras of Flying Co-Stereo within a unified world frame, we estimate the leader UAV's odometry\cite{geneva2020openvins} in the world frame while the follower estimates its relative pose to the leader. This leader-follower architecture offers distinct advantages over maintaining two independent VINS pose estimation systems, particularly in mitigating the accumulated drift of VINS pose estimation in the long term. In this section, the baseline (relative pose) estimation is transformed into the componental relative position and orientation estimation between two UAVs. 

\begin{figure}[]
  \centering
  \includegraphics[width=1.0\linewidth]{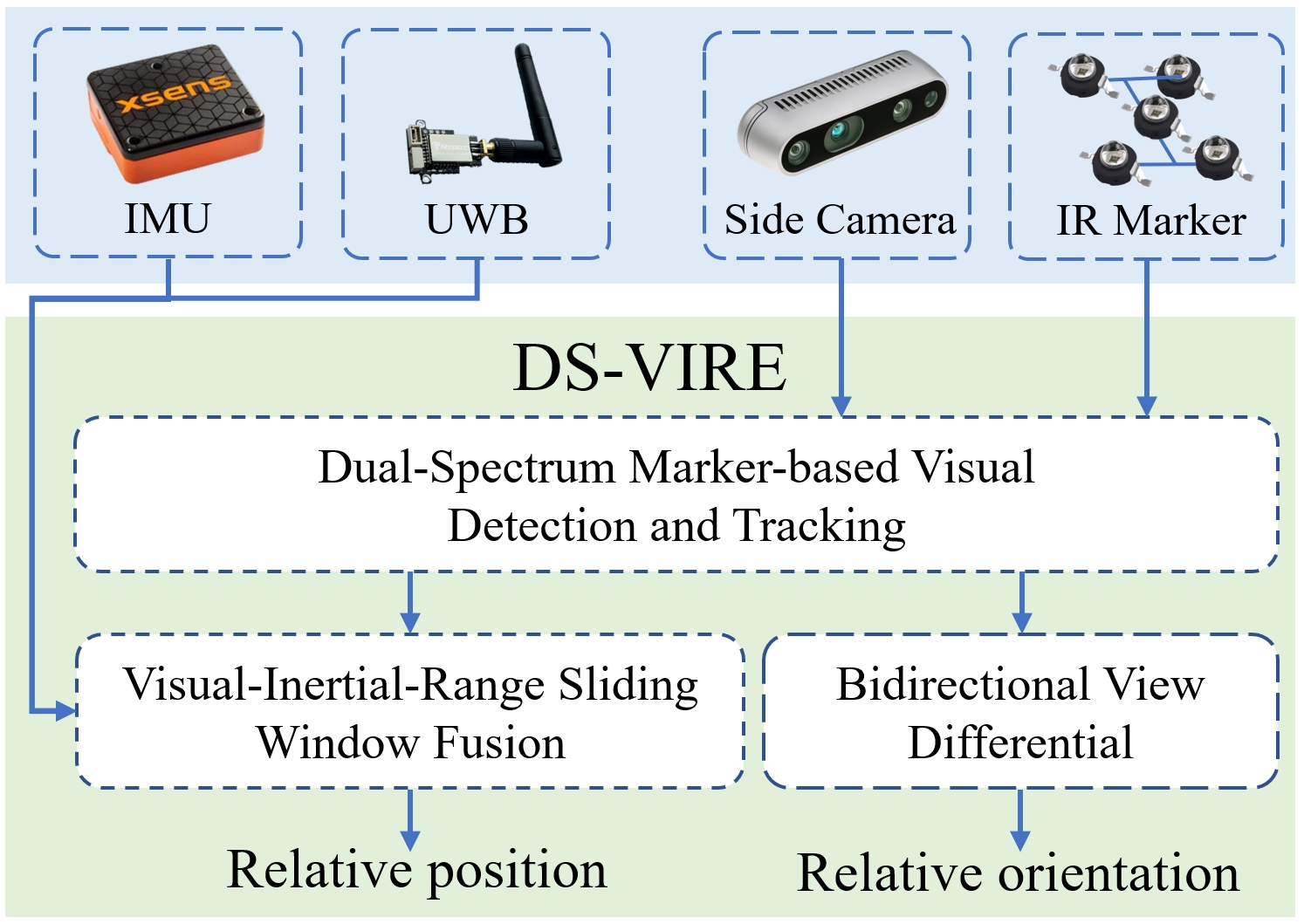}
  \caption{The framework of the DS-VIRE algorithm. The DS-MVDT submodule first detects and tracks the neighbor UAV's IR markers using the side camera. The visual observations are then fused with IMU and UWB ranging measurements within a sliding window to estimate the relative position. Finally, the BVD algorithm further bidirectionally differs from visual observation of the IR marker for precise relative orientation estimation.}
  \label{fig:2_DS_VIRE}
\end{figure}

We introduce the Dual-Spectrum Visual-Inertial-Ranging Estimator (DS-VIRE), illustrated in Fig.~\ref{fig:2_DS_VIRE}. The framework operates through three sequential components: (1) The Dual-Spectrum Marker-based Visual Detection and Tracking (DS-MVDT) algorithm, which employs side-mounted cameras to robustly detect and track infrared fiducial markers on neighboring UAVs; (2) A sliding window estimator fusing visual results of DS-MVDT with IMU and UWB measurements for relative position estimation; and (3) the Bidirectional View Differential (BVD) algorithm, which further processes the detected IR marker from DS-MVDT for precise relative orientation estimation.

\subsection{Dual-Spectrum Marker-based Visual Detection and Tracking}
\begin{figure}[]
    \centering
    \includegraphics[width=1.0\linewidth]{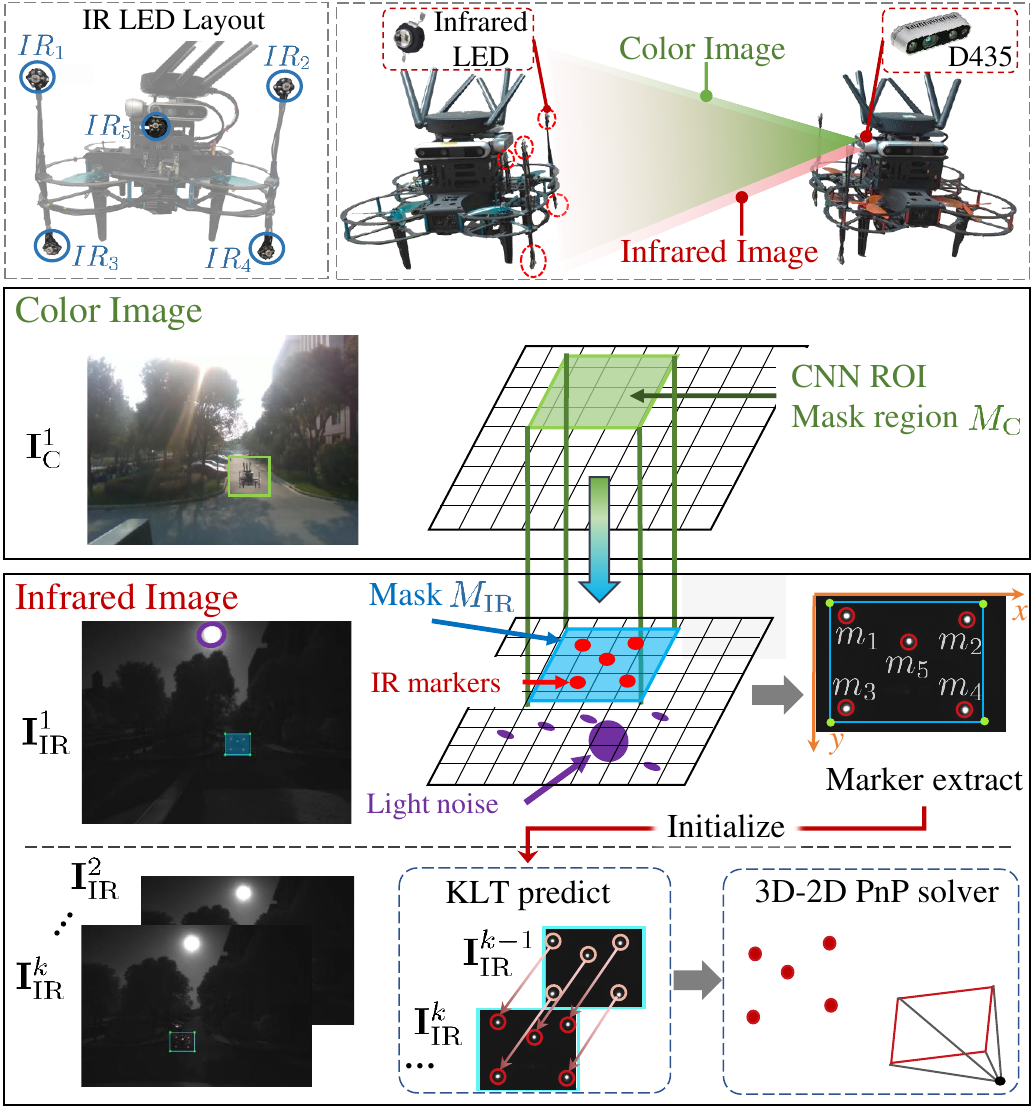}
    \caption{Pipeline of dual-spectrum marker-based visual detection and tracking.}
    \label{fig:3_DS_MVDT}
\end{figure}

The dual-spectrum marker-based visual detection and tracking algorithm (DS-MVDT) aims to provide robust and accurate visual observation of neighboring UAVs in outdoor environments.
The algorithm pipeline is presented in Fig.~\ref{fig:3_DS_MVDT}. The follower UAV1 is equipped with 850nm infrared LED markers, observed by the D435 camera side-mounted on the leader UAV. Notably, this observation is bidirectional, as the follower's side D435 camera similarly observes the leader's infrared markers. For illustrative purposes, we focus on the leader observing follower scenario.

To overcome the challenges of outdoor illumination variations and complex backgrounds, the dual-spectrum detection system capitalizes on the complementary characteristics of visible and infrared spectra. The visible spectrum, also referred to as the color image (hereafter briefly noted as `color image`), provides rich texture details that facilitate UAV identification but is vulnerable to interference from complex backgrounds. Conversely, the infrared image offers superior sensitivity to infrared light sources regardless of background texture, enabling robust marker detection while being susceptible to ambient light noise. The proposed dual-spectrum marker detection algorithm synergistically combines both imaging modalities.

The $\mathbf{I}^{k}_{\mathrm{C}}$ and $\mathbf{I}^{k}_{\mathrm{IR}}$ denote the $k$-th color and infrared images at timestamp $t_k$, respectively. The detection pipeline begins by processing the color image $\mathbf{I}^{1}_{\mathrm{C}}$ through a YOLOv4-tiny CNN \cite{bochkovskiy2020yolov4} to detect the follower UAV and generate a Region of Interest (ROI) as mask $M_\mathrm{C}$. The $M_\mathrm{C}$ is projected onto the aligned infrared image $\mathbf{I}^{1}_{\mathrm{IR}}$ by first normalizing with the color camera intrinsics, then transforming via the color-to-infrared extrinsics, and finally recovering with the infrared intrinsics to obtain the mask region $M_\mathrm{IR}$. The $M_\mathrm{IR}$ effectively filters infrared markers from ambient light interference, such as sunlight and reflective lights.

Within $M_\mathrm{IR}$, we employ FAST corner detection \cite{rosten2008FAST} with subpixel refinement to precisely locate fiducial markers in the infrared image $\mathbf{I}^{1}_{\mathrm{IR}}$. These markers, labeled as $m_i, i \in(1,2,3,4,5)$, correspond to a predefined spatial configuration. Then KLT tracker \cite{Lucas1981KLT} leverages optical flow to continuously track the marker blobs and iteratively predict their locations in subsequent infrared frames ($\mathbf{I}^{2}_{\mathrm{IR}}$ through $\mathbf{I}^{k}_{\mathrm{IR}}$), with geometric consistency checks ensuring alignment with the predefined marker layout. Meanwhile, the mask region $M_\mathrm{IR}$ adaptively updates its center and boundaries to enclose the predicted IR marker region, ensuring that optical flow computation remains unaffected by lighting disturbances.

Based on the infrared marker visual observations, we can estimate the marker-to-camera relative poses via PnP\cite{collins2014IPPEPnP}, which will be elaborated in Sec.~\ref{sec:vir_est}.   

To maximize computational efficiency, both CNN-based detection and infrared marker extraction are executed only during the initial phase and then suspended. The KLT tracker alone continuously tracks the marker blobs. When KLT tracking fails, the system reinitializes the marker points through reactivation of CNN detection and infrared extraction.

\subsection{Visual-Inertial-Ranging Sliding Window Estimation}\label{sec:vir_est}
We decouple the 6-DoF relative pose estimation between two UAVs into relative position and orientation estimations. 
For estimating the relative position $\mathbf{p}_{01}^k$, we fuse the visual feature observations from DS-MVDT, inertial, and ranging measurements within a sliding-window optimization framework. The estimation of relative orientation will be introduced in the subsequent Sec.~\ref{sec:bvd_est}. In the sliding-window optimization,   the relative position $\mathbf{p}_{01}^k$ and velocity $\mathbf{v}_{01}^k$ in the sliding window $\{k, k+1, ..., k+M-1\}$ are estimated:
\begin{equation}
  \begin{aligned}
  \mathcal{X}_{01} &= \{ \mathbf{p}_{01}^k, \mathbf{v}_{01}^k, \mathbf{p}_{01}^{k+1}, \mathbf{v}_{01}^{k+1},..., \mathbf{p}_{01}^{k+M-1}, \mathbf{v}_{01}^{k+M-1} \}
\end{aligned} 
\end{equation}
The cost function to be minimized in the sliding-window optimization is:
\begin{equation}
  \begin{array}{c}
    f(\mathcal{X}_{01}) = \mathop{\mathrm{argmin}}_{\mathcal{X}_{01}} \left\{ \sum\limits_{i=k}\limits^{k+M-1} \left\|\mathbf{r}_{\mathcal{V}}\left(\mathbf{p}_{\mathcal{V}_{01}}^i, \mathbf{p}_{\mathcal{V}_{10}}^i, \mathcal{X}_{01}\right)\right\|_{\mathbf{\Sigma}_{\mathcal{V}_{01}}^i}^2 \right. \\
    + \sum\limits_{i=k}\limits^{k+M-1} \left\|\mathbf{r}_{\mathcal{I}}\left(\mathbf{a}_{\mathcal{I}_{0}}^{i}, \mathbf{a}_{\mathcal{I}_{1}}^{i}, \mathcal{X}_{01}\right)\right\|_{\mathbf{\Sigma}_{\mathcal{I}_{01}}^{i}}^2 \\
    \left. + \sum\limits_{i=k}\limits^{k+M-1} \left\|\mathbf{r}_{\mathcal{R}}\left(d_{\mathcal{R}_{01}}^i, {\mathcal{X}}_{01}\right)\right\|_{\mathbf{\Sigma}_{\mathcal{R}_{01}}^i}^2 \right\}
    \end{array}
\end{equation}
where $\mathbf{r}_{\mathcal{V}}(\cdot), \mathbf{r}_{\mathcal{I}}(\cdot), \mathbf{r}_{\mathcal{R}}(\cdot)$ are the residuals constructed from Visual PnP, IMU, and UWB observations. $\mathbf{\Sigma}_{\mathcal{V}_{01}}^i, \mathbf{\Sigma}_{\mathcal{I}_{01}}^{i}, \mathbf{\Sigma}_{\mathcal{R}_{01}}^i$ are the covariance matrix of measurement noise. 

From the tracked visual feature observations of markers with DS-MVDT, we can utilize PnP\cite{collins2014IPPEPnP} to recover the relative poses between two UAVs based on known camera-marker and marker-body extrinsics. 
The mutual relative positions between UAV0 and UAV1 computed from PnP are denoted as $\tilde{\mathbf{p}}_{\mathcal{V}_{01}}^{i}$ and $\tilde{\mathbf{p}}_{\mathcal{V}_{10}}^{i}$, respectively. Then the visual residual in the sliding-window optimization at time $t_i$ can be formulated as follows:
\begin{equation}
  \mathbf{r}_{\mathcal{V}}\left(\mathbf{p}_{\mathcal{V}_{01}}^i, \mathbf{p}_{\mathcal{V}_{10}}^i, \mathcal{X}_{01}\right) = 
  \left[\begin{array}{c}
    \frac{1}{2}(\tilde{\mathbf{p}}_{\mathcal{V}_{01}}^{i} - {\mathbf{p}}_{01}^{i}) + \frac{1}{2} (\tilde{\mathbf{p}}_{\mathcal{V}_{10}}^{i} - {\mathbf{p}}_{01}^{i}) \\
	  \boldsymbol{0}
    \end{array}\right]
\end{equation}

The IMU integration is performed between consecutive timestamps $t_i$ and $t_{i+1}$ for both UAVs. The accelerations of the follower UAV are first transformed into the leader UAV’s frame using the relative rotation $\mathbf{R}_{01}^i$, which is assumed constant over the short interval $[t_i, t_{i+1}]$ following a standard first-order approximation: 

\begin{equation}
\begin{aligned}
\tilde{\mathbf{v}}_{\mathcal{I}_{01}}^{i+1} &= \mathbf{v}_{\mathcal{I}_{01}}^{i} + \int_{t=t_i}^{t_{i+1}} (\mathbf{R}_{01}^{i} \mathbf{a}_{\mathcal{I}_{1}}^{t} - \mathbf{a}_{\mathcal{I}_{0}}^{t}) dt 
\\
\tilde{\mathbf{p}}_{\mathcal{I}_{01}}^{i+1} &= \mathbf{p}_{\mathcal{I}_{01}}^i + \mathbf{v}_{\mathcal{I}_{01}}^{i}\Delta t + \int_{t_i}^{t_{i+1}} \int_{t_i}^{\tau} (\mathbf{R}_{01}^{i} \mathbf{a}_{\mathcal{I}_{1}}^{s} - \mathbf{a}_{\mathcal{I}_{0}}^{s}) ds d\tau
\end{aligned}
\end{equation}
The accelerometer biases are estimated online by each UAV's onboard OpenVINS system\cite{geneva2020openvins}, and we use the bias-compensated accelerations for computation, which is omitted here for brevity.
The residual of the IMU factor is calculated as follows:

\begin{equation}
  \mathbf{r}_{\mathcal{I}}\left(\mathbf{a}_{\mathcal{I}_{0}}^{i}, \mathbf{a}_{\mathcal{I}_{1}}^{i}, {\mathcal{X}}_{01}\right) = 
  \left[\begin{array}{c}
    \tilde{\mathbf{p}}_{\mathcal{I}_{01}}^{i+1} - {\mathbf{p}}_{01}^{i+1} \\
    \tilde{\mathbf{v}}_{\mathcal{I}_{01}}^{i+1}- {\mathbf{v}}_{01}^{i+1}
    \end{array}\right]
\end{equation}

The distance measurement by UWB between UAV0 and UAV1 is $d_{\mathcal{R}_{01}}^i$, and the residual of the UWB factor is as follows:
\begin{equation}
  \mathbf{r}_{\mathcal{R}}\left(d_{\mathcal{R}_{01}}^i, {\mathcal{X}}_{01}\right) = 
  \left[\begin{array}{c}
    d_{\mathcal{R}_{01}}^{i} -  \left\| {\mathbf{p}}_{{01}}^{i}\right\|\\
	\boldsymbol{0}
    \end{array}\right]
\end{equation}
Finally, the cost function $f(\mathcal{X}_{01})$ is soloved by the Levenberg-Marquardt algorithm from Ceres-solver\cite{agarwal2012ceres}.

\subsection{Bidirectional View Differential Module}\label{sec:bvd_est}
We now discuss the estimation of the relative orientation between two UAVs. 
The roll and pitch angles $(\theta_0, \phi_0, \theta_1, \phi_1)$ measured by IMU0 on UAV0 and IMU1 on UAV1 are represented in gravity-aligned coordinate frames and are generally accurate. For simplicity, we use the IMU frames interchangeably with the UAV body frames. Thus, the relative roll and pitch can be directly computed as $\theta_{01} = \theta_1 - \theta_0$ and $\phi_{01} = \phi_1 - \phi_0$, respectively. However, the yaw angles $(\psi_0, \psi_1)$ measured by the IMUs on the two UAVs exhibit significant deviations and are generally less reliable.

\begin{figure}[]
  \centering
  \includegraphics[width=1.0\linewidth]{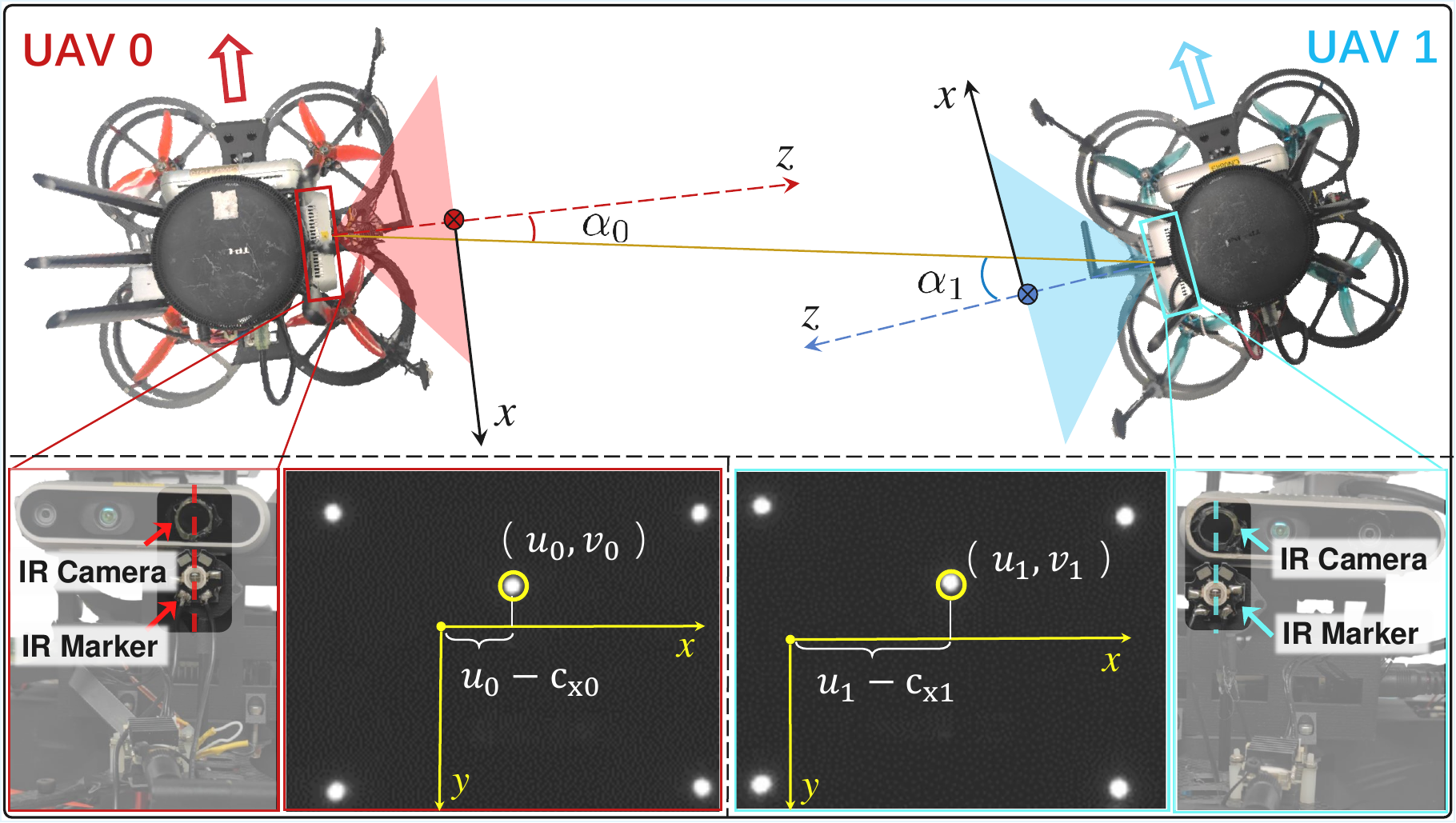}
  \caption{Illustration of bidirectional visual differential for relative yaw estimation.}
  \label{fig:4_BVD}
\end{figure}

Due to the long distance between the camera and the IR markers, as well as the limited size of the markers mounted on the UAVs, the orientation estimation derived from visual PnP—based on feature associations provided by DS-MVDT—is often noisy (as demonstrated in later experiments). To address this, we propose a collaborative Bidirectional View Differential (BVD) method to estimate the relative yaw $\psi_{01}$ between UAVs. 
The key insight is to compute the relative yaw by differencing the bearing measurements of the side-view camera's centers on two UAVs (depicted in Fig.~\ref{fig:4_BVD}). 
Detecting the camera center in images is challenging; therefore, we use a bright IR marker as a proxy of the camera center. 
Figure~\ref{fig:4_BVD} shows that the D435 IR camera on UAV0 observes the IR marker on UAV1 with bearing $\alpha_0$, while symmetrically, UAV1's camera observes UAV0's marker at $\alpha_1$. A close-up view in Fig.~\ref{fig:4_BVD} shows the proximity between the D435 IR camera and the central IR marker on each UAV's side panel. This configuration allows the marker position to serve as an accurate proxy in the horizontal plane or in the bird's view. 
Clearly, under near-hovering conditions where the UAVs' roll and pitch angles are close to zero, the relative yaw between UAV0 and UAV1 can be estimated from their perspective angles as $\psi_{01} = \alpha_1 - \alpha_0$. Each perspective angle $\alpha$ can be derived from the pixel location of the center IR marker. Taking UAV0 as an example, given the pixel coordinates $(u_0, v_0)$, the view angle $\alpha_0$ is computed as $\alpha_0 = \mathrm{arctan}((u_0 - c_{x0})/f_{x0})$, where $f_{x0},c_{x0}$ are the focal length and principal point in the x-axis. Thus, the relative yaw $\psi_{01}$ is given by:

\begin{equation}\label{eq:yaw}
  \psi_{01} = \mathrm{arctan}((u_1 - c_{x1})/f_{x1}) - \mathrm{arctan}((u_0 - c_{x0})/f_{x0})
\end{equation}

When roll and pitch deviate from zero, we convert the 3D position of the center IR marker (obtained from the DS-MVDT module’s visual PnP results) from the current camera pose to an equivalent 3D position under a level camera pose (roll = pitch = 0).
The corrected 3D position is then projected onto the image plane, and the resulting pixel location is used to compute the relative yaw $\psi_{01}$ following~\eqref{eq:yaw}.

\section{Cross-camera Feature Association}
This section introduces the detection of co-visible landmarks and the real-time, persistent cross-camera feature association within the overlapping stereo view of Flying Co-Stereo. These components form the foundation for subsequent collaborative triangulation. 
SuperPoint\cite{detone2018superpoint} and SuperGlue\cite{sarlin2020superglue} are employed for landmark detection and cross-camera matching due to their superior accuracy. However, despite leveraging CUDA and TensorRT for acceleration, their high computational demands impose a significant burden on onboard processing, resulting in noticeable latency and constraining the frequency of feature association. To alleviate this, we propose a hybrid strategy that periodically performs SuperPoint detection and SuperGlue cross-camera matching on sparsely sampled keyframes, combined with continuous temporal feature tracking within each individual camera.

Unlike conventional methods that rely solely on SuperPoint and SuperGlue for inter-UAV feature matching, our approach leverages periodic SuperPoint-SuperGlue associations as guided feature correspondences. These guided features are then continuously predicted across new image frames in real time within each camera view. Meanwhile, the continuous feature tracking and inheritance accumulate sufficient observations for landmarks for subsequent robust triangulation. Additionally, despite the continuous change in viewpoints, the inter-frame variation within each UAV's field of view is minimal, making it particularly suitable for individual feature tracking. We refer to this lightweight and efficient framework as Guidance-Prediction SuperPoint-SuperGlue (GP-SS). 

\begin{figure}[]
  \centering
  \includegraphics[width=1.0\linewidth]{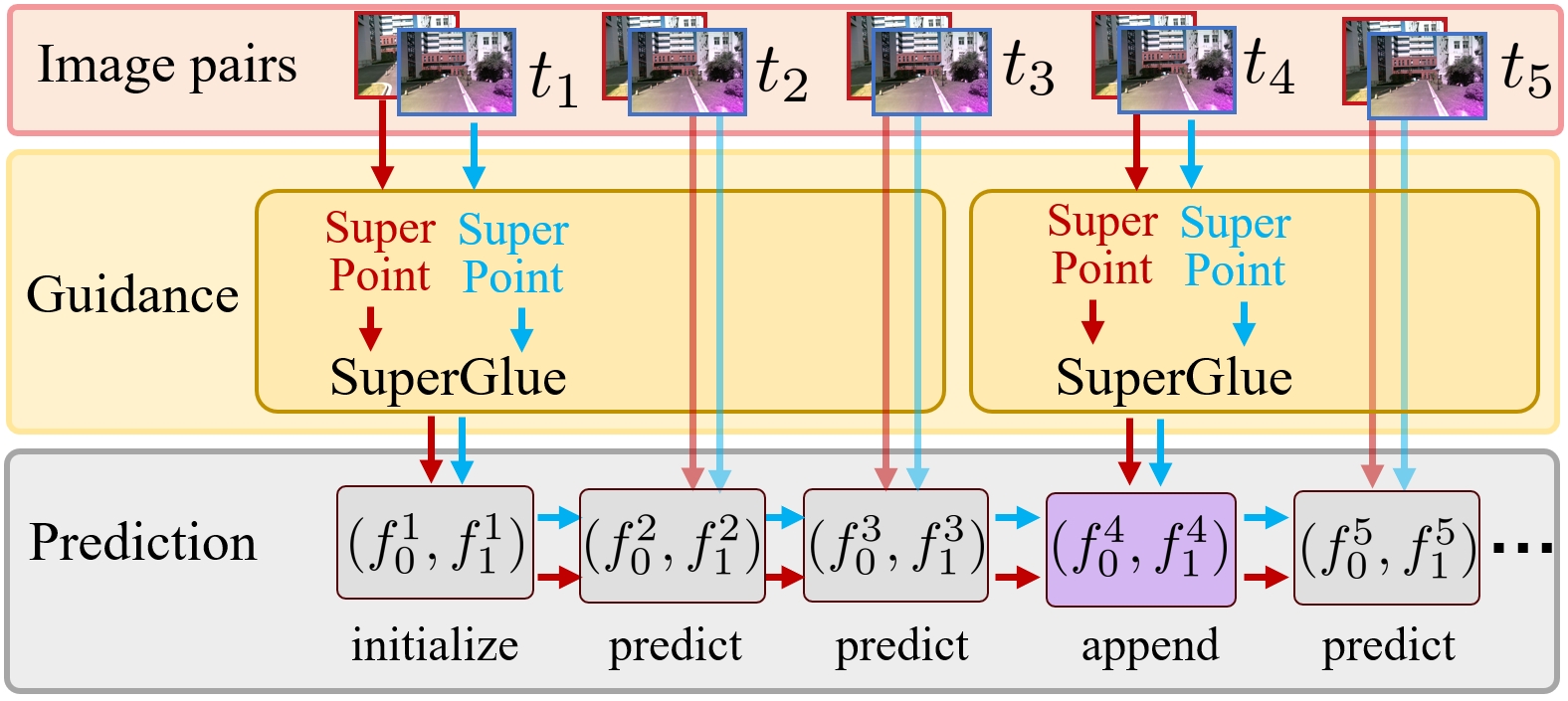}
  \caption{The diagram of guidance-prediction cross-camera feature association.}
  \label{fig:5_feature_diagram}
\end{figure}

\begin{figure}[]
  \centering
  \includegraphics[width=1.0\linewidth]{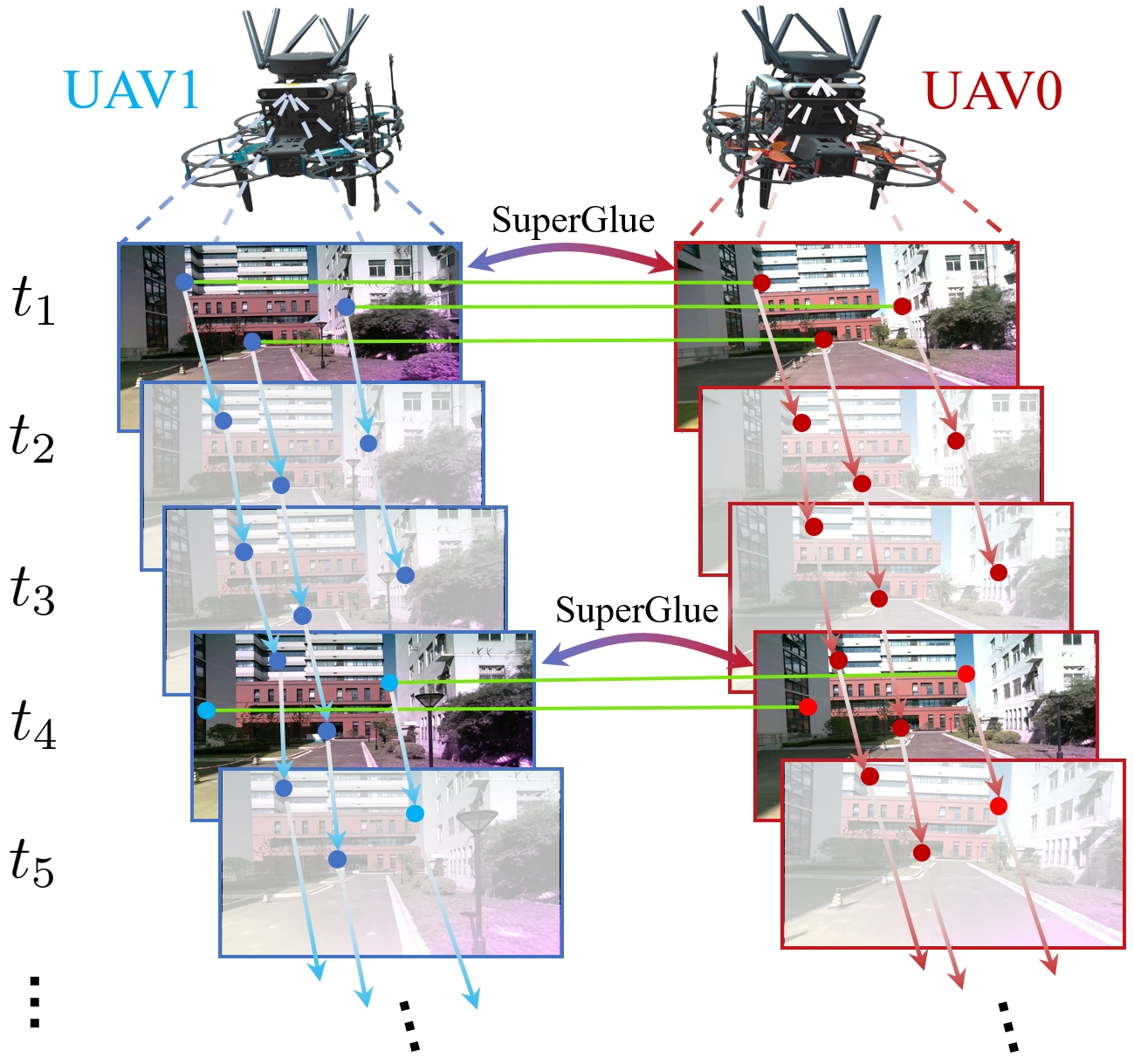}
  \caption{The periodic matching guidance and continuous feature prediction.}
  \label{fig:6_feature_illustration}
\end{figure}

The guidance-prediction architecture is illustrated in Fig.~\ref{fig:5_feature_diagram}. Timestamped images from UAV0 are paired with the closest-timestamped images from UAV1. Let $f_m^k$ denote the feature from UAV $m, m\in(0,1)$ at the $k$-th frame. SuperPoint features extracted from local views are matched via SuperGlue to establish initial associations ($f_0^1 \leftrightarrow f_1^1$). Subsequently, each UAV tracks its features over time using the Lucas-Kanade optical flow (LK-flow), predicting ($f_0^2, f_1^2$), ($f_0^3, f_1^3$), etc., in the following image pairs. Since $f_0^2$ inherits the ID of $f_0^1$ and $f_1^2$ inherits the ID of $f_1^1$, persistent cross-camera associations (e.g., $f_0^2 \leftrightarrow f_1^2$) are maintained based on these inherited IDs. This allows the system to accumulate cross-view 2D observations of co-visible landmarks across consecutive frames, until tracking fails in either view, as shown in Fig.~\ref{fig:6_feature_illustration}.

In our implementation, SuperPoint and SuperGlue require approximately 74 ms per inference on the NVIDIA NX platform, resulting in the skipping of two frames in a 30 Hz image stream. The second round of cross-camera matching via SuperGlue occurs at timestamp $t_4$, where a new set of guided features is appended to the previously predicted features ($f_0^3, f_1^3$), resulting in updated associations ($f_0^4, f_1^4$). The prediction process then continues on subsequent images at timestamp $t_5$.

In summary, the guidance-prediction design leverages SuperPoint and SuperGlue as periodic guidance for cross-camera feature association. Meanwhile, continuous feature prediction via LK-flow ensures real-time correspondence inheritance, enabling the accumulation of a sufficient number of persistent cross-view feature observations between the two UAVs.

\section{Large-scale Sparse to Dense Mapping}
Building upon the camera poses obtained from online baseline estimation and the associated 2D observations of co-visible landmarks from both UAVs, this section first performs collaborative triangulation to estimate the 3D positions of co-visible sparse landmarks over a long range. Subsequently, a monocular depth network is applied to UAV0’s images to generate scale-ambiguous but dense depth estimates. Finally, the sparse landmarks are exponentially fitted with the monocular depth to predict a large-scale dense map.

\subsection{Collaborative Triangulation of Sparse Landmarks}
This section presents the collaborative triangulation of co-visible landmarks. Accurate and robust landmark triangulation requires both sufficient feature observations and adequate parallax. For distant landmarks along the heading direction, the parallax induced by the forward motion of a single UAV is often insufficient. In contrast, the Flying Co-Stereo system, leveraging a wide-baseline binocular configuration during forward flight, provides ample parallax for improved triangulation accuracy.

Consider a landmark $\mathbf{p}_f$ observed from 1 to $k$ timestamps by the camera of UAV0 (frame $C_0^1$ to $C_0^k$) and the camera of UAV1 (frame $C_1^1$ to $C_1^k$). The landmark $\mathbf{p}_f$ in two cameras at $k$ timestamp are represented as ${}^{C_0^k}\mathbf{p}_f, {}^{C_1^k}\mathbf{p}_f$. For the collaborative triangulation, we select the first camera frame of UAV0 as the anchor frame ${{}^{A}(\cdot)}$. The $\mathbf{p}_f$ in the anchor frame is also represented as ${}^{A}\mathbf{p}_f$. The transformation of feature representations from other frames of UAV0 and UAV1, spanning timestamps 1 to $k$, into this anchor frame is expressed as follows:

\begin{equation}
  ^A\mathbf{p}_f = ^A\mathbf{R}_{C_m^i} {}^{C_m^i}\mathbf{p}_f + ^A\mathbf{p}_{C_m^i}, m = 0,1, i = 1,2,...,k
\end{equation}
where the camera pose of UAV0 and UAV1 w.r.t. anchor frame are denoted as ${}^{A}\mathbf{p}_{C_0^i},{}^{A}\mathbf{R}_{C_0^i}, {}^{A}\mathbf{p}_{C_1^i}, {}^{A}\mathbf{R}_{C_1^i}$, respectively. The ${}^{C_m^i}\mathbf{p}_f$ can be further formulated as the depth ${}^{C_m^i}z_f$ and bearing vector ${}^{C_m^i}\mathbf{b}_f$ and represented in anchor frame.

\begin{equation}
\begin{aligned}
  ^A\mathbf{p}_f & = ^A\mathbf{R}_{C_m^i} ^{C_m^i}z_f ^{C_m^i} \mathbf{b}_f + {}^A \mathbf{p}_{C_m^i} \\
  & =^{C_m^i}z_f {}^A\mathbf{b}_{C_m^i \rightarrow f} + {}^A\mathbf{p}_{C_m^i}
  \end{aligned}
\end{equation}

Then we remove the degree of freedom of depth $^{C_m^i}z_f$ by multiplying the orthogonal space $^A \mathbf{N}_m^i = \left\lfloor{ }^A \mathbf{b}_{C_m^i \rightarrow f} \times\right\rfloor$ to the bearing vector. We can get the following equation:

\begin{equation}
\begin{aligned}
  ^A \mathbf{N}_m^i {}^A \mathbf{p}_f & = {}^A \mathbf{N}_m^i {}^{C_m^i}z_f {}^A \mathbf{b}_{C_m^i \rightarrow f} + {}^A \mathbf{N}_m^i {}^A \mathbf{p}_{C_m^i} \\
  & = {}^A \mathbf{N}_m^i {}^A \mathbf{p}_{C_m^i}
  \end{aligned}
\end{equation}

We stack all the measurements of UAV0 and UAV1 from $1$ to $k$ timestamps.

\begin{equation}
\underbrace{\left[\begin{array}{c}
  {}^A \mathbf{N}_0^1 \\
  \vdots \\
  {}^A \mathbf{N}_0^k \\
  {}^A \mathbf{N}_1^1 \\
  \vdots \\
  {}^A \mathbf{N}_1^k \\
  \end{array}\right]}_{\mathbf{A}_{6k\times3}} {}^A \mathbf{p}_f = \underbrace{\left[\begin{array}{c}
  {}^A \mathbf{N}_0^1 {}^A \mathbf{p}_{C_0^1} \\
  \vdots \\
  {}^A \mathbf{N}_0^k {}^A \mathbf{p}_{C_0^k} \\
  {}^A \mathbf{N}_1^1 {}^A \mathbf{p}_{C_1^1} \\
  \vdots \\
  {}^A \mathbf{N}_1^k {}^A \mathbf{p}_{C_1^k} \\
  \end{array}\right]}_{\mathbf{b}_{6k\times1}}
  \label{eq:triangulation}
\end{equation}

The quick solution of $^A\mathbf{p}_f$ can be obtained from:

\begin{equation}
\mathbf{A}_{6k\times3}^{\top} \mathbf{A}_{6k\times3}  {}^A\mathbf{p}_f = \mathbf{A}_{6k\times3}^{\top} \mathbf{b}_{6k\times1}
\label{eq:condition_number}
\end{equation}

Finally, we validate the triangulated ${}^A \mathbf{p}_f$ by evaluating the condition number of the matrix $\mathbf{A}^{\top} \mathbf{A}$. Any ${}^A\mathbf{p}_f$ with a condition number exceeding a predefined threshold is discarded. When additional observations from $C_0^{k+1}$ and $C_1^{k+1}$ are incorporated into the sliding window, the ${}^A\mathbf{p}_f$ is further refined through Gauss-Newton optimization by minimizing the reprojection error across all views in the window.

The aforementioned co-visible landmarks, constrained by the overlapping stereo fields of view, are typically located in distant, jointly observed regions. To enhance spatial diversity, near-field landmarks are also incorporated, primarily sourced from the onboard VIO system—OpenVINS \cite{geneva2020openvins} in our setup.

\subsection{Exponential Dense Depth Prediction}

We begin by analyzing the spatial distribution characteristics of the aforementioned sparse landmarks, followed by the introduction of our sparse-to-dense depth prediction framework based on exponential fitting.

To ensure computational efficiency, the Flying Co-Stereo system maintains only 60–200 valid sparse 3D landmarks, corresponding to just 0.03\% pixel density for a 640×480 image. This is significantly lower than methods like VOID\cite{wong2020VOID} and KBNet\cite{wong2021unsupervisedKBNet}, which sample around 1500 points (0.5\% density) from dense maps to simulate VIO landmarks. As a result, our setting falls into the ultra-sparse to dense reconstruction regime, where conventional sparse-to-dense completion methods are not directly applicable.

The most similar approach to ours is AB-VINS \cite{merrill2024ABVINS}, which performs a linear fitting of VIO landmarks and monocular depth estimates using scale (A) and bias (B) corrections. This method relies on the affine-invariant monocular depth distribution property of MiDaS\cite{ranftl2020MiDas}. However, the affine-invariant property tends to hold only at close range, and the predicted depth distribution can deviate significantly in long-range, especially large-scale outdoor environments.

\begin{figure}[]
  \centering
  \includegraphics[width=1.0\linewidth]{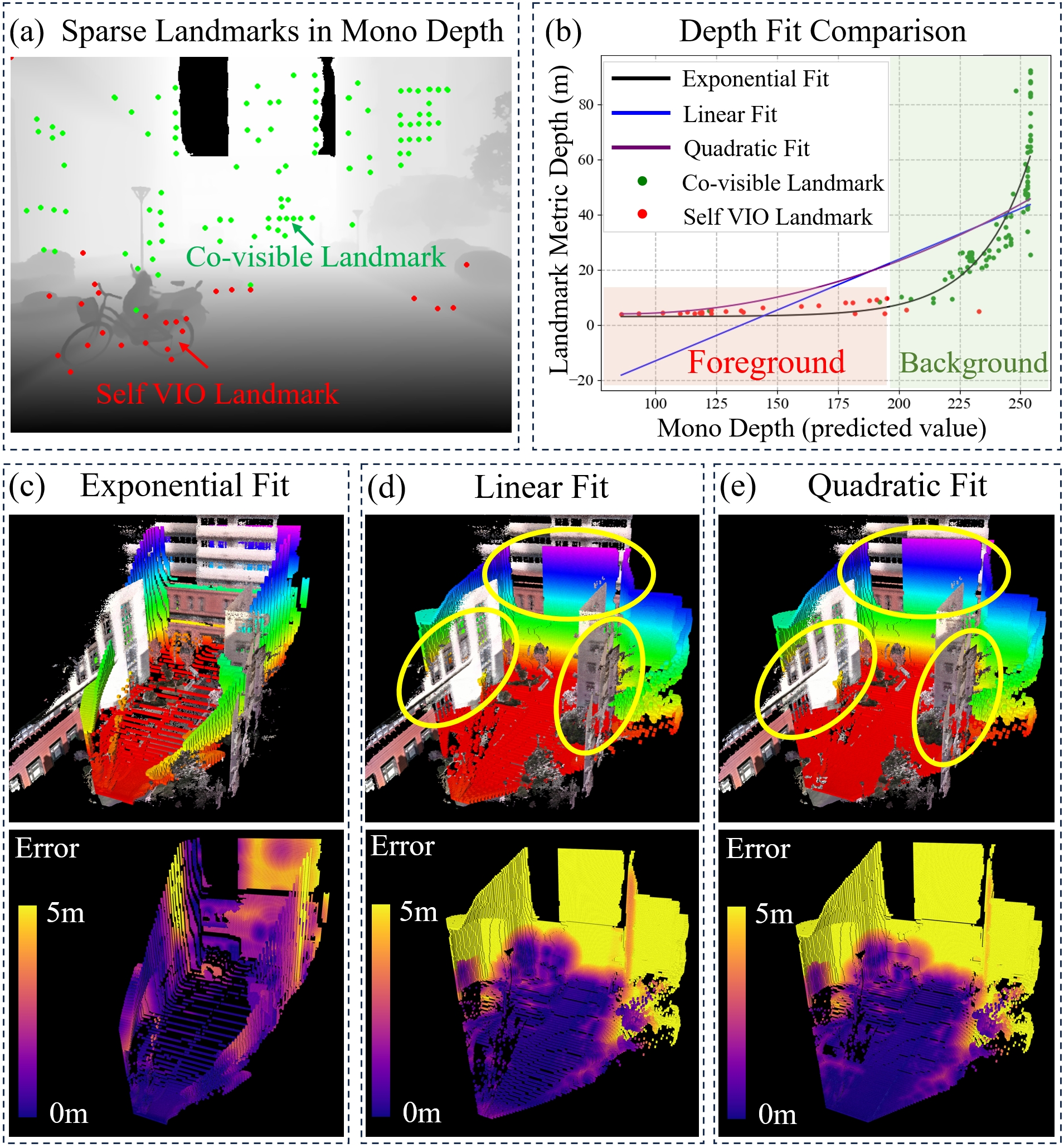}
  \caption{Comparison of different fitting methods for dense reconstruction. (a) Sparse landmarks overlaid on monocular depth: red for UAV0’s VIO landmarks, green for co-visible landmarks. (b) Predicted monocular depth versus triangulated metric depth for sparse landmarks, with exponential, linear, and quadratic fitting curves. (c)-(e) Dense point clouds generated by each fitting method. Yellow circles highlight regions with depth underestimation. The error heatmap shows the depth error distribution.}
  \label{fig:7_depth_fit}
\end{figure}

We carefully analyze the spatial distribution of landmarks and monocular depth estimates from the perspective of UAV0. As illustrated in Fig.~\ref{fig:7_depth_fit}(a), the sparse landmarks include UAV0’s VIO landmarks (red) and co-visible landmarks (green). VIO landmarks mainly cluster in nearby areas, while the co-visible landmarks typically distribute in distant co-observed regions. The depth fit in Fig.~\ref{fig:7_depth_fit}(b) visualizes the predicted up-to-scale depth value by DepthAnythingV2\cite{yang2024DepthAnythingV2} of landmarks with their metric depth measurements. The red VIO landmarks cluster in the foreground, exhibiting a linear depth progression, while the green co-visible landmarks are distributed in the background, demonstrating an exponential growth pattern. The near-field distribution supports the use of linear fitting in AB-VINS. In contrast, the exponential trend in the long range is attributed to the inherent difficulty for monocular depth networks like DepthAnythingV2 to maintain accuracy across large depth disparities. Based on the depth distribution pattern, we propose an exponential fitting approach to recover dense metric depth from monocular depth estimates and sparse landmarks.

The exponential fitting model is designed to align the monocular depth estimates with the actual metric depths of the sparse 3D landmarks, enabling accurate scale refinement. At timestamp $k$, we obtain $N$ 3D landmarks denoted as ${}^{C_0^k}\mathbf{p}_f^n$, $n = 1,2,..., N$. For notational simplicity, we omit the frame superscript and express each landmark as $\mathbf{p}_f^n = [x_f^n, y_f^n, z_f^n]^{\top}$. Then, a monocular depth network DepthAnythingV2\cite{yang2024DepthAnythingV2} is employed to predict the depth image $\mathbf{D}$. We extract the depth component $z_f^n$ from each triangulated landmark $\mathbf{p}_f^n$, and perform an exponential fitting against the corresponding predicted monocular depth value $d_f^n$ from the monocular depth image $\mathbf{D}$. The loss function of fitting is formulated as follows:

\begin{equation}
  \mathop{\mathrm{min}}_{a,b,c,d}  \sum\limits_{n=1}\limits^{N} (z_f^n - a \cdot \exp(b \cdot (d_f^n - c)) + d) 
\end{equation}
with $a, b, c, d$ as the parameters. The parameters are estimated by minimizing a loss function using the Ceres Solver\cite{agarwal2012ceres} with analytical Jacobians, after which the optimized $a, b, c, d$ are applied to the entire predicted depths of the monocular depth image to recover the metrically scaled dense depth.

This exponential formulation effectively handles both the linear fitting requirements for close-range landmarks and the exponential growth characteristics of long-range landmarks. The comparative analysis of exponential, linear, and quadratic fitting methods is presented in Fig.~\ref{fig:7_depth_fit}(c)-(e), where the exponential fitting yields more accurate map reconstruction compared to the linear and quadratic alternatives.

For the downstream applications, the multi-frame dense metric depth prediction can be further fused using Truncated Signed Distance Function (TSDF) fusion, generating large-scale Euclidean Signed Distance Fields (ESDF) \cite{oleynikova2017voxblox}. These large-scale ESDF representations enable early path planning and ensure safe navigation in environments containing large-scale obstacles such as buildings or terrain.

\section{Experiments and Discussion}
\subsection{Single-UAV System Configuration}

\begin{figure}[]
  \centering
  \includegraphics[width=1.0\linewidth]{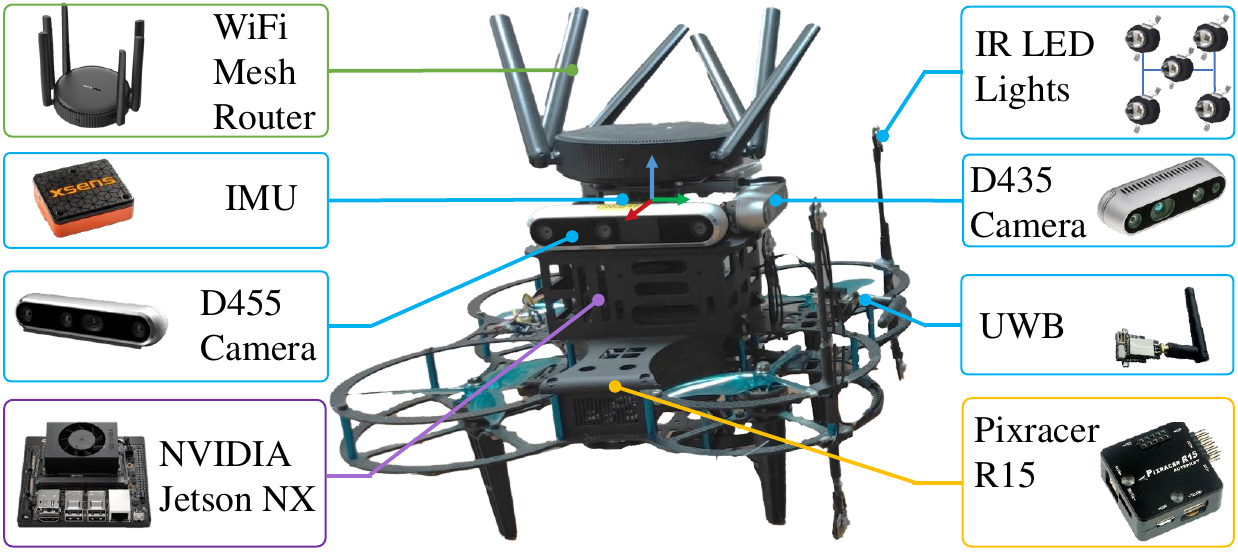}
  \caption{The hardware layout of the custom-developed quadcopter UAV.}
  \label{fig:sensor_layout}
\end{figure}

The Flying Co-Stereo adopts two custom-developed quadcopter UAVs with the same configuration. As shown in Fig.~\ref{fig:sensor_layout}, each UAV is equipped with an Intel RealSense D455, whose forward-facing color camera supports collaborative mapping. The D455’s built-in infrared stereo cameras serve as a benchmark for conventional stereo perception. An Xsens MTi-630 IMU is mounted at the UAV center, and its measurements, together with the D455 color images, are processed by OpenVINS to estimate the UAV’s pose in the world frame. 
The Intel RealSense D435 is mounted on the side of the UAV for mutual observation of fiducial IR markers. The fiducial IR marker consists of five infrared lights (850 nm wavelength) arranged to provide robust visual cues for estimating the relative pose between the two UAVs.
The Nooploop UWB radio (model LinkTrack LTPS) is placed on the side of the UAV for range measurement between two UAVs.  A WiFi mesh router (model TP-Link WDR7651) is placed on the top of the UAV to construct a local mesh network between two UAVs, supporting high-bandwidth and low-latency data exchange. An NVIDIA Jetson Xavier NX (8G) \footnote{\url{https://www.nvidia.cn/autonomous-machines/embedded-systems/jetson-xavier-nx/}} is adopted as the onboard computer. All programs are developed by the Robot Operating System (ROS) in the Linux Ubuntu 20.04 system and coded in C++14. The flight controller adopts Pixracer R15 with customized PX4\footnote{\url{https://github.com/PX4/PX4-Autopilot}} firmware. The UAV's takeoff weight is approximately 1.6 kg with a 16.8V 4000-mAh battery, offering a maximum endurance of about 7 minutes, which suffices for our tests.

\subsection{Inter-UAV Communication and Coordination}
\subsubsection{\textbf{Inter-UAV Communication}}
\begin{figure}[]
  \centering
  \includegraphics[width=1.0\linewidth]{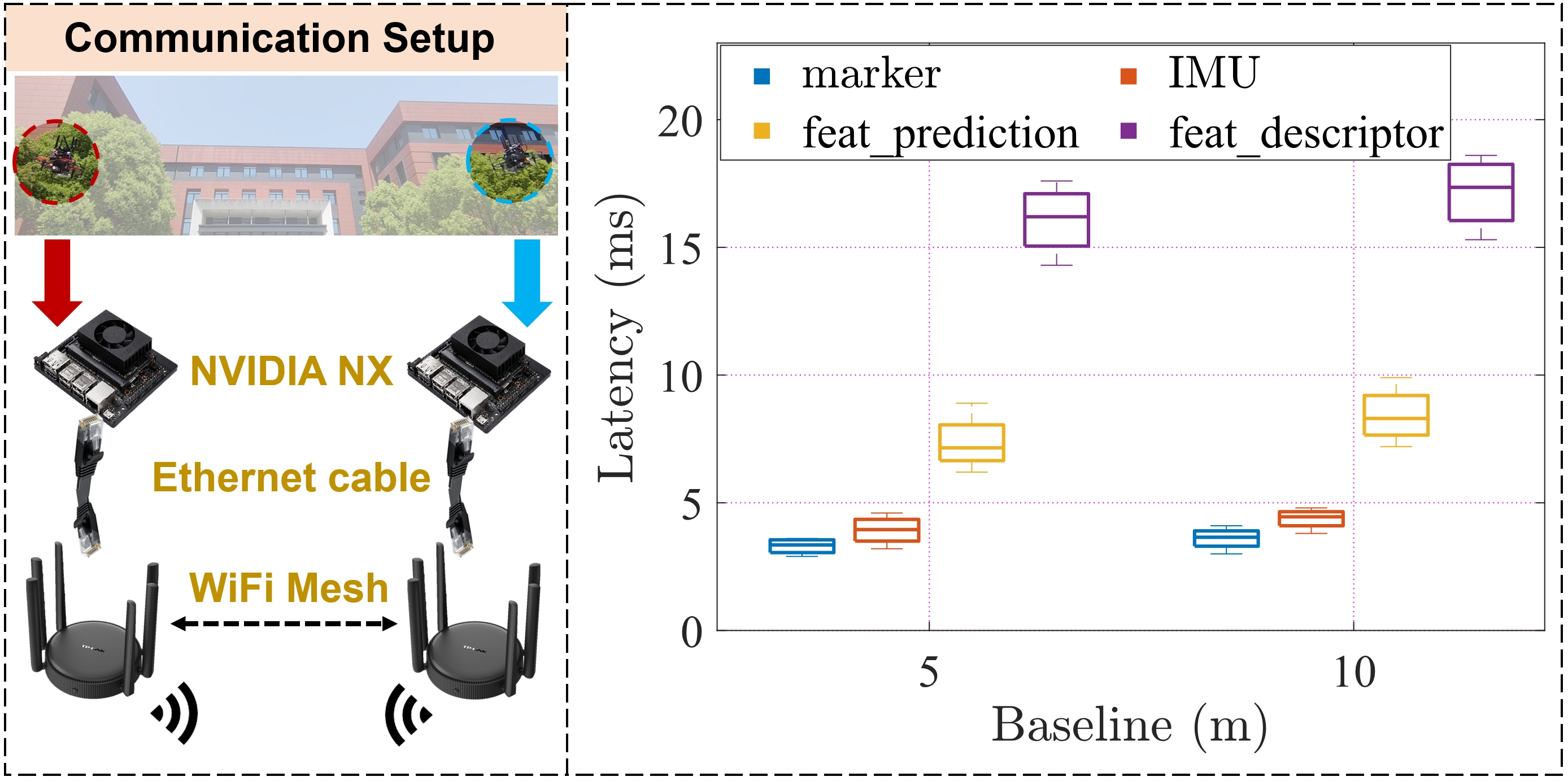}
  \caption{The latency of meshed WiFi communication with different baselines.}
  \label{fig:latency}
\end{figure}

As illustrated in Fig.~\ref{fig:latency}, each UAV's onboard computer interfaces with a WiFi router via Gigabit Ethernet, enabling efficient data exchange through the meshed WiFi network. Two UAVs are connected in the same LAN (Local Area Network), where the data is coded by LCM (Lightweight Communications and Marshalling)\footnote{\url{https://lcm-proj.github.io/lcm/}} and exchanged with an efficient UDP protocol.

The two UAVs exchange two types of data: pose estimation and feature association. The total bandwidth is approximately 705KB/s, comprising: (1) infrared marker observations (8B timestamp + 8B pixel coordinates) at 30 Hz, (2) IMU data (8B timestamp + 12B acceleration + 16B quaternion) at 30 Hz, (3) 150 SuperPoint pixels for prediction (150 $\times$ (2B + 8B)) at 30 Hz, and (4) 50 SuperPoint descriptors for guidance and matched feature IDs (50 $\times$ (259 $\times$ 4B + 2B)) at 13 Hz.

Experiments show that data transmission latency increases with packet size. For a 5 m unobstructed baseline, the average latencies are 3.2 ms (markers), 3.8 ms (IMU), 7.2 ms (feature predictions), and 16 ms (feature descriptors). The 16 ms latency is acceptable given the SuperGlue update interval of 74 ms at 13 Hz. Latency remains stable over longer baselines. In our setup, the two UAVs fly in parallel within a 5 m range in open environments, allowing the mesh WiFi routers to maintain consistently low-latency communication.

We acknowledge that the Flying Co-Stereo system currently operates under well-established communication, pausing mapping during interruptions and resuming upon reconnection.

\subsubsection{\textbf{Asynchronization handling}}
We carefully handle the time asynchrony between two UAVs. First, we achieve system clock synchronization within 1 ms precision using the Network Time Protocol (NTP). Second, hardware-triggered synchronous exposure for cameras on the two UAVs remains technically challenging. Therefore, we implement a software-based synchronization, which involves nearest-neighbor pairing of exposure timestamps between both UAVs. The online baseline estimation from the DS-VIRE also undergoes interpolation for camera exposure time, employing linear interpolation for translation and spherical linear interpolation (SLERP) for rotation.

\subsubsection{\textbf{Coordinated flight}}
We briefly describe the coordinated flight policy between the two UAVs. The leader UAV0 performs path planning and follows the trajectory using its VIO system. The follower UAV1 estimates the relative pose in UAV0’s frame via the DS-VIRE module and actively controls its motion to maintain the desired relative pose in real time.

\subsubsection{\textbf{Computation Distribution}}
SuperPoint features are extracted locally on each UAV, while UAV0 handles cross-agent matching, triangulation, and dense reconstruction. Online baseline estimation is performed on UAV1 to support feedback control.

\subsection{Dynamic Baseline Estimation}
The dynamic baseline estimation of Flying Co-Stereo, which corresponds to the relative pose estimation between two UAVs, is evaluated in terms of both accuracy and robustness.
Specifically, the accuracy is assessed indoors using a motion capture system, while the robustness is validated through outdoor experiments.

\begin{figure}[]
  \centering
  \includegraphics[width=1.0\linewidth]{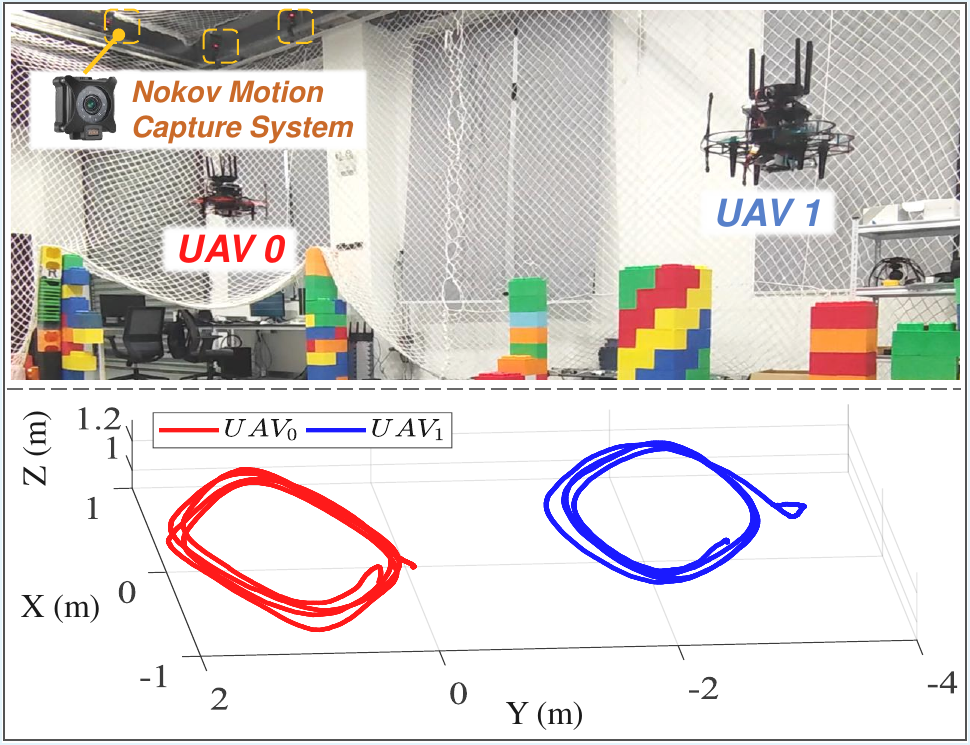}
  \caption{The experiments for relative pose estimation of Flying Co-Stereo under the motion capture system.}
  \label{fig:experiments_rel_pose_setup}
\end{figure}

\subsubsection{\textbf{Accuracy Evaluation of Baseline Estimation}}
The indoor accuracy evaluation is performed using a NOKOV motion capture system (Fig.~\ref{fig:experiments_rel_pose_setup}), where both UAVs autonomously fly synchronized circular trajectories in the ENU coordinate frame (x-forward, y-left, z-up) with a baseline of approximately 3 m. DS-VIRE provides relative position and orientation estimation. For comparison, we evaluate a visual PnP-based method using only inter-UAV observations, and a VIO differencing method \cite{weinstein2018visual}, which derives the relative pose by subtracting independently estimated VIO poses from each UAV.

For initialization, both DS-VIRE and the visual PnP method support convenient mutual observation-based initialization, while the VIO differencing method requires external motion capture to align the separate VIO coordinate frames.

\begin{figure*}[]
  \centering
  \includegraphics[width=1.0\linewidth]{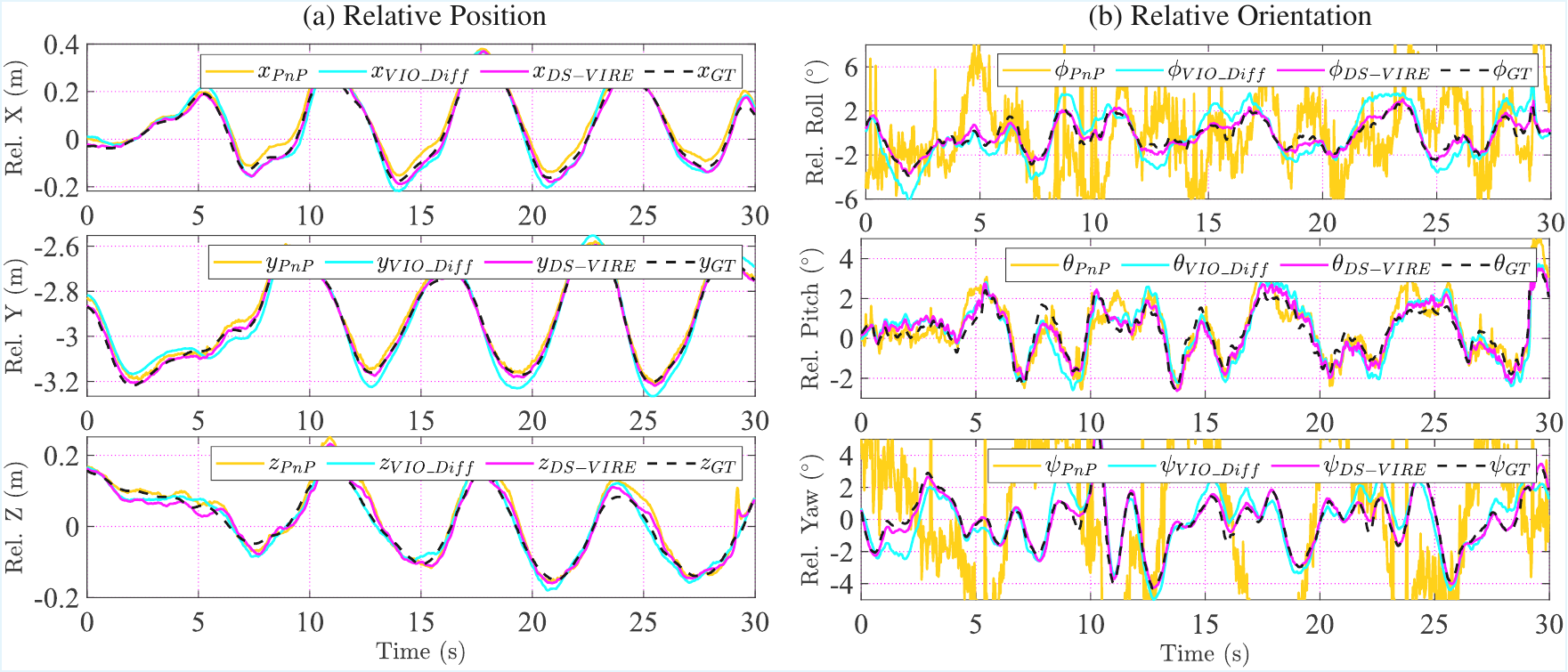}
  \caption{Comparison of relative position and orientation estimates from the proposed DS-VIRE, Visual PnP, and VIO\_Diff methods with ground truth.}
  \label{fig:exp_rel_pose}
\end{figure*}

The estimated relative position is presented in Fig.~\ref{fig:exp_rel_pose}(a), with the corresponding error analysis summarized in Table~\ref{tab:pos_error}. The total MAE (mean absolute error considering all x/y/z directions) of relative position estimation using DS-VIRE is 0.013 m, significantly outperforming the visual PnP-based method and VIO differencing method, which yield 0.018 m and 0.024 m, respectively.

\begin{table}[]
  \caption{Relative Position Error Analysis}
  \label{tab:pos_error}
  \setlength{\tabcolsep}{1.4mm}{
  \begin{tabular}{ccccccccc}
  \toprule
  \multirow{2}{*}{\begin{tabular}[c]{@{}c@{}}\textbf{Method} \end{tabular}}  & \multicolumn{4}{c}{\textbf{MAE (m)}}   & \multicolumn{4}{c}{\textbf{RMSE (m)}}  \\ \cline{2-9} 
            & Total & $x$ & $y$ & $z$   & Total & $x$ & $y$ & $z$ \\ \hline
  DS-VIRE         & \textbf{0.013}  &  \textbf{0.013}      & \textbf{0.009}     & 0.015    & \textbf{0.028} & \textbf{0.016}  & \textbf{0.011}  & 0.019        \\ 
  Visual PnP     & 0.018  &  0.016      & 0.014     & 0.020    & 0.036 & 0.032  & 0.025  & 0.026      \\ 
  VIO\_Diff     & 0.024  &  0.022      & 0.034     & \textbf{0.014}    & 0.047 & 0.026  & 0.039  & \textbf{0.018}      \\ 
  \bottomrule      
  \end{tabular}
  }
\end{table}

\begin{table}[]
  \caption{Relative Orientation Error Analysis}
  \label{tab:ori_error}
  \setlength{\tabcolsep}{1.4mm}{
  \begin{tabular}{ccccccccc}
  \toprule
  \multirow{2}{*}{\begin{tabular}[c]{@{}c@{}}\textbf{Method} \end{tabular}}  & \multicolumn{4}{c}{\textbf{MAE ($^\circ$)}}  & \multicolumn{4}{c}{\textbf{RMSE ($^\circ$)}}  \\ \cline{2-9} 
            & Total & $\phi$ & $\theta$ & $\psi$   & Total & $\phi$ & $\theta$ & $\psi$ \\ \hline
  DS-VIRE       & \textbf{0.286}  &  \textbf{0.303}  & \textbf{0.340}     & \textbf{0.214}    & \textbf{0.619} & \textbf{0.369}  & \textbf{0.401}  & \textbf{0.292}   \\ 
  Visual PnP     & 2.689  &  2.905      & 0.614     & 4.549    & 6.656 & 3.616  & 0.772  & 5.534      \\ 
  VIO\_Diff      & 0.739  &  0.975      & 0.528     & 0.715     & 1.599  & 1.188   & 0.616  & 0.874      \\ 
  \bottomrule      
  \end{tabular}
  }
\end{table}

This improvement is mainly attributed to DS-VIRE's persistent visual pose constraints, which serve as a stable baseline, further enhanced by the smoothing properties of IMU integration and the absolute range constraints from UWB. In contrast, the VIO differencing method is more susceptible to system drift, resulting in less accurate relative position estimates.

The relative orientation estimation is presented in Fig.~\ref{fig:exp_rel_pose}(b). The orientation error analysis, expressed in terms of roll ($\phi$), pitch ($\theta$), and yaw ($\psi$), is summarized in Table~\ref{tab:ori_error}. The BVD module in DS-VIRE leverages bidirectional visual differential observations of IR markers, achieving high-precision yaw estimation with MAE of 0.214${}^\circ$. In contrast, the visual PnP-based method suffers from noisy rotation estimates due to the compact IR marker layout and long observation distance. In-plane rotations induce only subtle pixel-level variations, leading to high estimation uncertainty. The VIO differencing approach is still affected by VIO drifts of two independent UAV systems.

We further evaluate the system performance across baseline distances ranging from 2 m to 5 m. The total position and orientation errors (considering all x/y/z directions) are presented in Table~\ref{tab:pose_err_diff_baselines}. The decline in position estimation accuracy with increasing baseline is mainly due to amplified observation errors in marker-based visual PnP. In contrast, orientation estimation remains stable, primarily benefiting from IMU differencing and bidirectional visual differential, which are less sensitive to long-range baselines.

\begin{table}[]
  \caption{Relative Pose Error Analysis with Different Baselines}
  \label{tab:pose_err_diff_baselines}
  \setlength{\tabcolsep}{1.7mm}{
  \begin{tabular}{cccccc}
  \toprule
  \multirow{2}{*}{\begin{tabular}[c]{@{}c@{}}\textbf{Method} \end{tabular}} & \multirow{2}{*}{\begin{tabular}[c]{@{}c@{}}\textbf{Baseline} \end{tabular}} & \multicolumn{2}{c}{\textbf{Position}}   & \multicolumn{2}{c}{\textbf{Orientation}}  \\ \cline{3-6} 
              & & \multicolumn{1}{c}{MAE (m)} & \multicolumn{1}{c}{RMSE (m)} & \multicolumn{1}{c}{MAE ($^\circ$)} & \multicolumn{1}{c}{RMSE($^\circ$)} \\ \hline
              \multirow{4}{*}{DS-VIRE}  & 2m      &  0.011      & 0.026     & 0.281    & 0.610   \\ 
                                      & 3m      &  0.013      & 0.028     & 0.286    & 0.619   \\ 
                                      & 4m      &  0.018      & 0.037     & 0.289    & 0.621   \\ 
                                      & 5m      &  0.025      & 0.052     & 0.290    & 0.625   \\ \bottomrule      
  \end{tabular}
  }
\end{table}

\begin{figure*}[]
  \centering
  \includegraphics[width=1.0\linewidth]{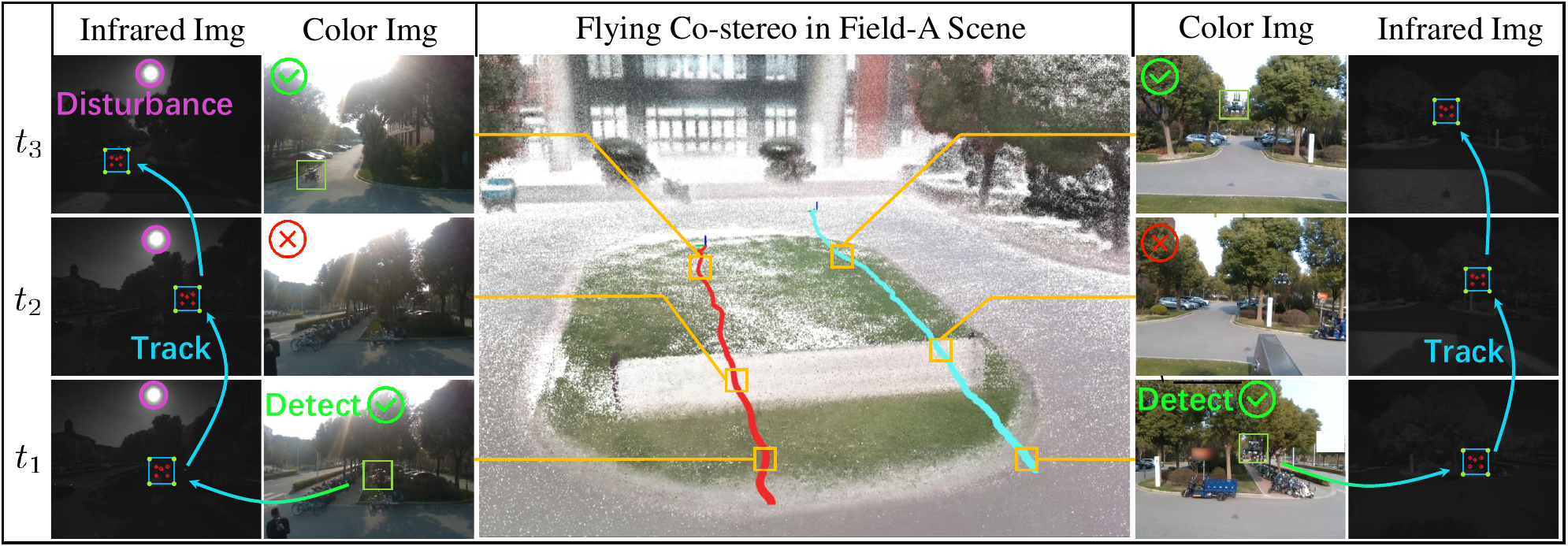}
  \caption{Robustness evaluation of DS-MVDT in outdoor Field-A, featuring complex backgrounds and lighting disturbances. UAV trajectories are shown in red and blue.}
  \label{fig:exp_rel_robustness}
\end{figure*}

\subsubsection{\textbf{Robustness Evaluation of Baseline Estimation}}
The robustness of DS-VIRE is essential for ensuring reliable relative pose estimation in complex outdoor environments. This robustness largely stems from visual observations, which are handled by the Dual-Spectrum Marker-based Visual Detection and Tracking (DS-MVDT) module. In outdoor scenarios, visual interference arises from cluttered backgrounds, varying lighting conditions, and light-induced noise. To evaluate system robustness under diverse environmental conditions, we conduct experiments across multiple outdoor scenes.

As shown in Fig.~\ref{fig:exp_rel_robustness}, flight experiments are conducted in outdoor Field-A. Focusing on the red trajectory, DS-MVDT initially detects the UAV using a region of interest (ROI) in the color image, which is projected onto the infrared image to form a mask that suppresses sunlight interference and enables reliable IR marker extraction. Although YOLOv4-tiny fails midway due to background clutter, DS-MVDT continues tracking the IR markers in the infrared stream. The mask region dynamically updates its center and boundaries to enclose the predicted marker area, further enhancing robustness. A similar process is applied to the UAV on the blue trajectory.

\begin{table}[]
  % \centering
  \caption{Success Rate Comparison of Visual Observation Methods Across Diverse Scenes}
  \label{tab:vdt_rate}
  \setlength{\tabcolsep}{1.3mm}
  \begin{tabular}{ccccc}
  \toprule
  \textbf{Datasets} & \textbf{Description} & \begin{tabular}[c]{@{}c@{}}\textbf{Total} \\ \textbf{frame}\end{tabular} & \begin{tabular}[c]{@{}c@{}} \textbf{VDT} \\ \textbf{frame (ratio)} \end{tabular} & \begin{tabular}[c]{@{}c@{}} \textbf{DS-MVDT} \\ \textbf{frame (ratio)} \end{tabular} \\ 
  \hline 
  Field-A & Against the sunlight & 628 & 109 (17\%) & 608 (96\%) \\

  Field-B  & Intense  sunlight & 770 & 336 (43\%) & 762 (99\%)  \\ 
  
  Field-C & Cluttered background & 1112 & 781 (70\%) & 1110 (99\%) \\

  Field-D  & Noisy lights & 756 & 168 (22\%) & 734 (97\%) \\

  Field-E & Remote observation & 795 & 510 (64\%) & 786 (98\%) \\   \bottomrule

  \end{tabular}
\end{table}

\begin{figure}[]
  \centering
  \includegraphics[width=1.0\linewidth]{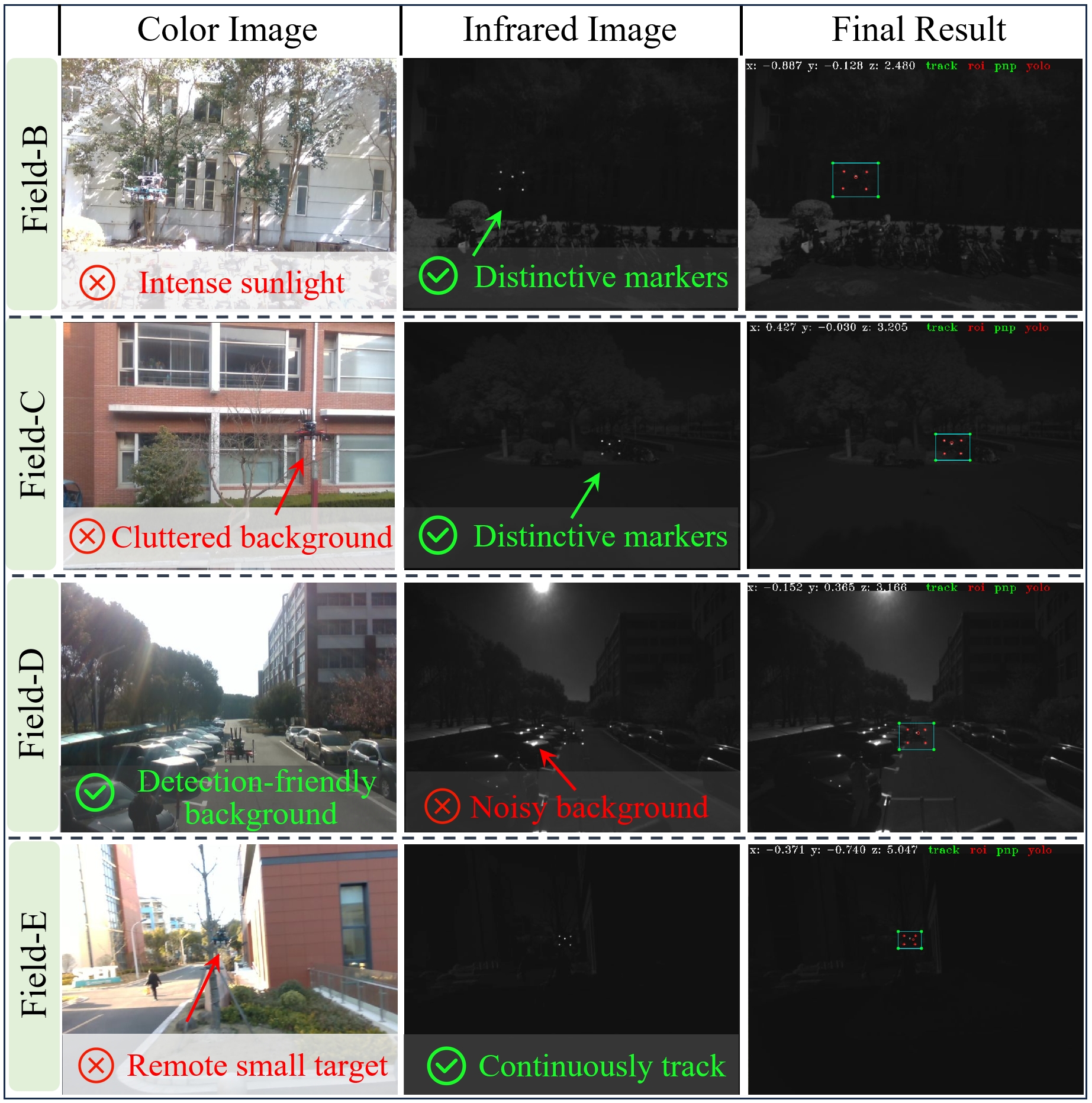}
  \caption{Experiments of DS-MVDT with challenges from intense sunlight, cluttered background, light noises, and remote observation.}
  \label{fig:exp_rel_robustness_multi_scenes}
\end{figure}

We further conduct comprehensive field experiments across four challenging outdoor environments (Field-B to Field-E) to rigorously evaluate the DS-MVDT algorithm's robustness. As illustrated in Fig.~\ref{fig:exp_rel_robustness_multi_scenes}, each scenario presents unique challenges: intense sunlight interference, complex background clutter, severe infrared noise contamination, and long observation distances. Despite these challenging conditions, the DS-MVDT algorithm consistently delivers superior performance through three key mechanisms: (1) effective suppression of lighting disturbances via ROI extraction from the color image, (2) robust infrared marker detection leveraging the infrared spectrum's insensitivity to cluttered backgrounds, and (3) adaptive infrared marker tracking for robust target following. The system successfully maintains accurate UAV tracking, proving its robustness in real-world scenarios.

We adopt the Visual Detection and Tracking (VDT) module from Omni-Swarm \cite{xu2022omni} as a baseline, which combines YOLOv4-tiny for UAV detection with MOSSE tracking \cite{bolme2010MOSSE}, relying on successful CNN redetection to maintain tracking. We evaluate DS-MVDT against VDT in the same real-world scenarios by comparing the ratio of successfully tracked frames. As shown in Table~\ref{tab:vdt_rate}, DS-MVDT consistently achieves over 96\% tracking success across all test environments, while VDT exhibits variable performance (17–70\%) due to its sensitivity to environmental conditions. These results highlight DS-MVDT’s robustness in maintaining continuous tracking without the need for frequent CNN redetection.

\subsection{Cross-Camera Feature Association}

\begin{figure}[]
  \centering
  \includegraphics[width=1.0\linewidth]{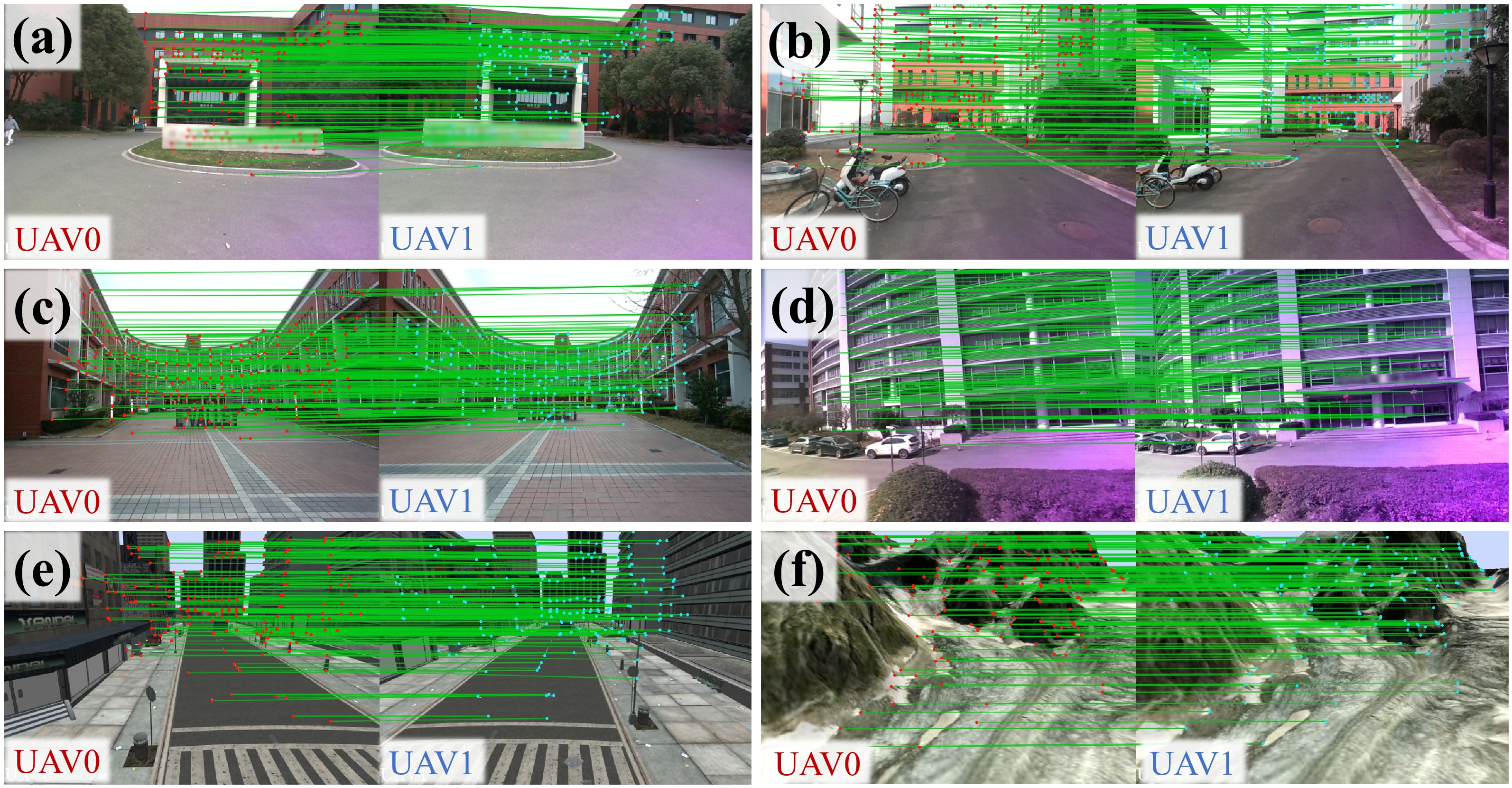}
  \caption{The visualization of cross-camera feature association in the real-world and simulated scenes. (a)-(d) show the feature association in real-world scenarios (Field-A to Field-D). (e)-(f) present the feature association in the simulated city and mountain scenes, respectively.}
  \label{fig:exp_feature_match_visual}
\end{figure}

Our evaluation first examines the real-time performance of cross-camera feature association with run-time analysis, followed by an analysis of the temporal persistence of co-visible features across consecutive frames and their cumulative observation statistics. We conduct extensive experiments across diverse real-world environments and additionally incorporate simulated environments to further enrich the variety of testing scenarios. Fig.~\ref{fig:exp_feature_match_visual} presents the cross-camera feature matching results between UAV0 and UAV1.

\subsubsection{\textbf{Run-Time Analysis}}
In real-world experiments, the Intel RealSense D455 is utilized as the forward-facing camera, providing 640$\times$480 color images at 30 Hz. To ensure computational efficiency on NVIDIA NX, we leverage the JetPack module (CUDA 11.4 and TensorRT 8.4) to accelerate SuperPoint and SuperGlue network inference, while LK-flow computations are handled by OpenCV with CUDA acceleration.

We compare the proposed Guidance-Prediction SuperPoint with SuperGlue (GP-SS) with several popular feature association methods: the original SuperPoint with SuperGlue (SS), ORB\cite{murartal_orb-slam2_2017} with nearest neighbor matching\cite{david2004SIFT} and RANSAC filtering\cite{RANSAC1981} (ORB\_NR), and SURF\cite{bay2006surf} with nearest neighbor matching and RANSAC filtering (SURF\_NR). The experimental results presented in Table~\ref{tab:feature_frequency} demonstrate that GP-SS achieves near-30Hz feature association frequencies across all test scenarios, significantly outperforming the baseline SS method at 13Hz while also showing superior performance compared to both SURF and ORB-based approaches.

\begin{table}[]
  \centering
  \caption{Frequency Analysis of Cross-camera Feature on Self-constructed Datasets}
  \label{tab:feature_frequency}
  \setlength{\tabcolsep}{3.4mm}{
  \begin{tabular}{ccccc}
  \toprule
  Datasets    & GP-SS      & SS   & SURF\_{NR}  & ORB\_{NR}  \\ \hline
  Field-A    & \textbf{30} hz    & 13 hz & 7 hz & 13 hz\\
  Field-B    & \textbf{29} hz    & 12 hz & 6 hz & 13 hz\\
  Field-C    & \textbf{28} hz     & 12 hz    & 6 hz     & 12 hz   \\
  Field-D    & \textbf{29} hz     & 13 hz    & 6 hz     & 12 hz    \\
  \bottomrule
  \end{tabular}
  }
\end{table}

\begin{table}[]
  \centering
  \caption{Run-time Analysis on Self-constructed Datasets}
  \label{tab:feature_run_time}
  \setlength{\tabcolsep}{1.8mm}{
  \begin{tabular}{ccccc}
  \toprule
  Modules    & GP-SS      & SS   & SURF\_NR  & ORB\_NR  \\ \hline
  Feature Detect         & 38.0 ms    & 38.0 ms & 126.5 ms & 35.3 ms\\
  Feature Match          & 37.0 ms    & 37.0 ms & 31.4 ms & 40.7 ms\\
  LK-flow       & 7.4 ms     & -    & -     & -    \\
  Total Run-Time & \textbf{7.4/82.4} ms & 75.0 ms & 157.9 ms & 76.0 ms\\
  \bottomrule
  \end{tabular}
  }
\end{table}

\begin{figure}[]
  \centering
  \includegraphics[width=1.0\linewidth]{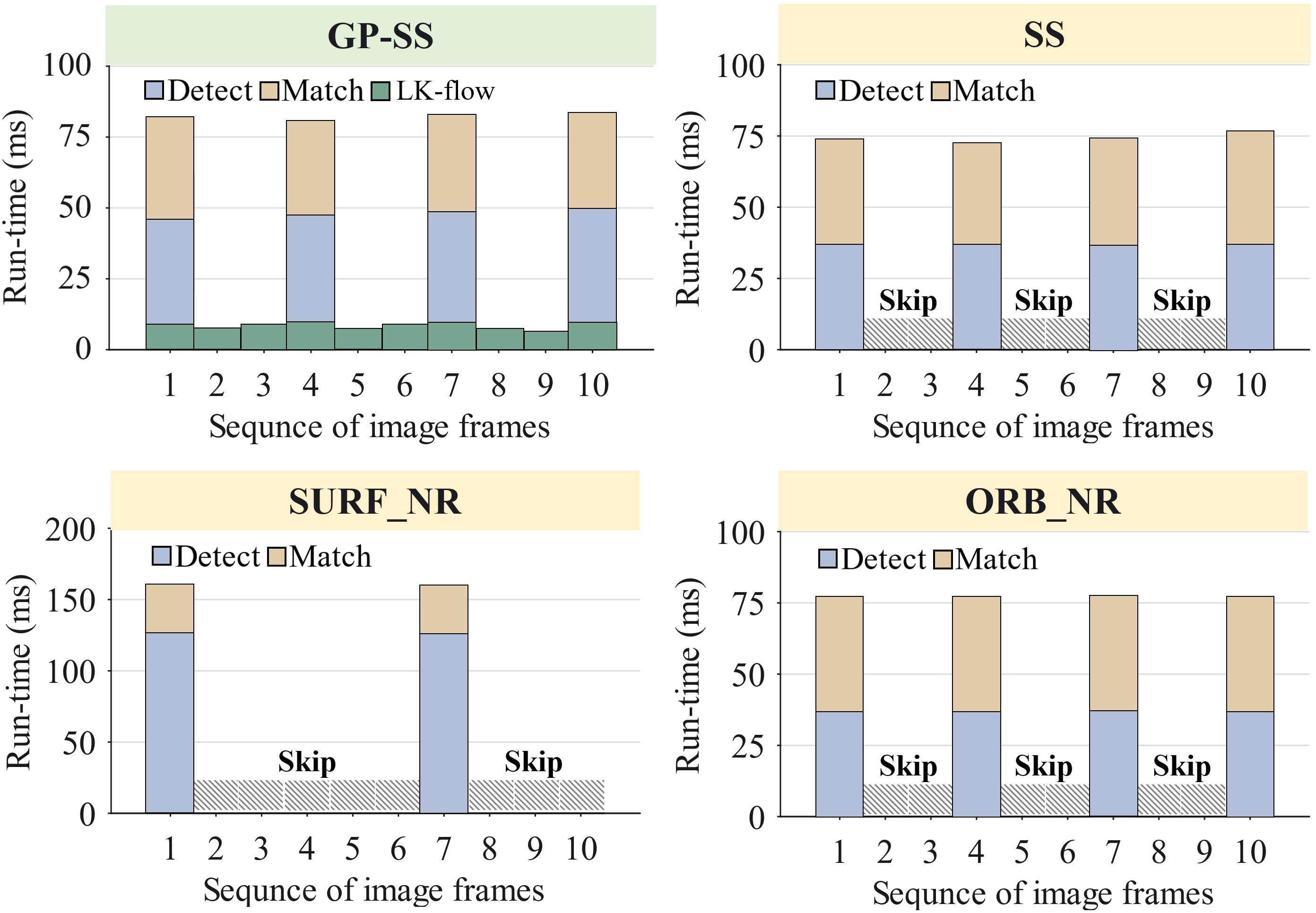}
  \caption{The run-time of different cross-camera feature association algorithms along the image sequence.}
  \label{fig:exp_feature_time_cost}
\end{figure}

The average time consumption is detailed in Table~\ref{tab:feature_run_time}. While wide-parallax cross-agent feature detection and matching are computationally expensive, feature prediction via LK-flow is lightweight and efficient. Leveraging this characteristic, the proposed GP-SS adopts a guidance-prediction strategy: periodic SuperPoint-SuperGlue matches serve as guidance, and the guided features are then predicted to subsequent frames in real time using optical flow. As illustrated in Fig.~\ref{fig:exp_feature_time_cost}, GP-SS effectively utilizes the intermediate frames between SuperPoint-SuperGlue executions to continuously generate new feature matches, significantly increasing the frequency and efficiency of feature associations.

In comparison, the baseline SS method can only generate new matches periodically. Due to the high computational cost of cross-camera matching, it skips several intermediate frames after each matching cycle, resulting in inefficient frame utilization and a lower frequency of feature association. The SURF\_NR and ORB\_NR methods show similar timing characteristics. 

\subsubsection{\textbf{Temporal Persistence of Features}} We further evaluate feature persistence across consecutive frames within each camera view to ensure sufficient multi-frame observations. In GP-SS, intra-camera feature association is achieved via LK-flow tracking. We select AirVO \cite{xu2023airvo} as comparison, which uses SuperPoint for new feature detection and SuperGlue for frame-to-frame matching and inheritance.

\begin{figure}[]
  \centering
  \includegraphics[width=1.0\linewidth]{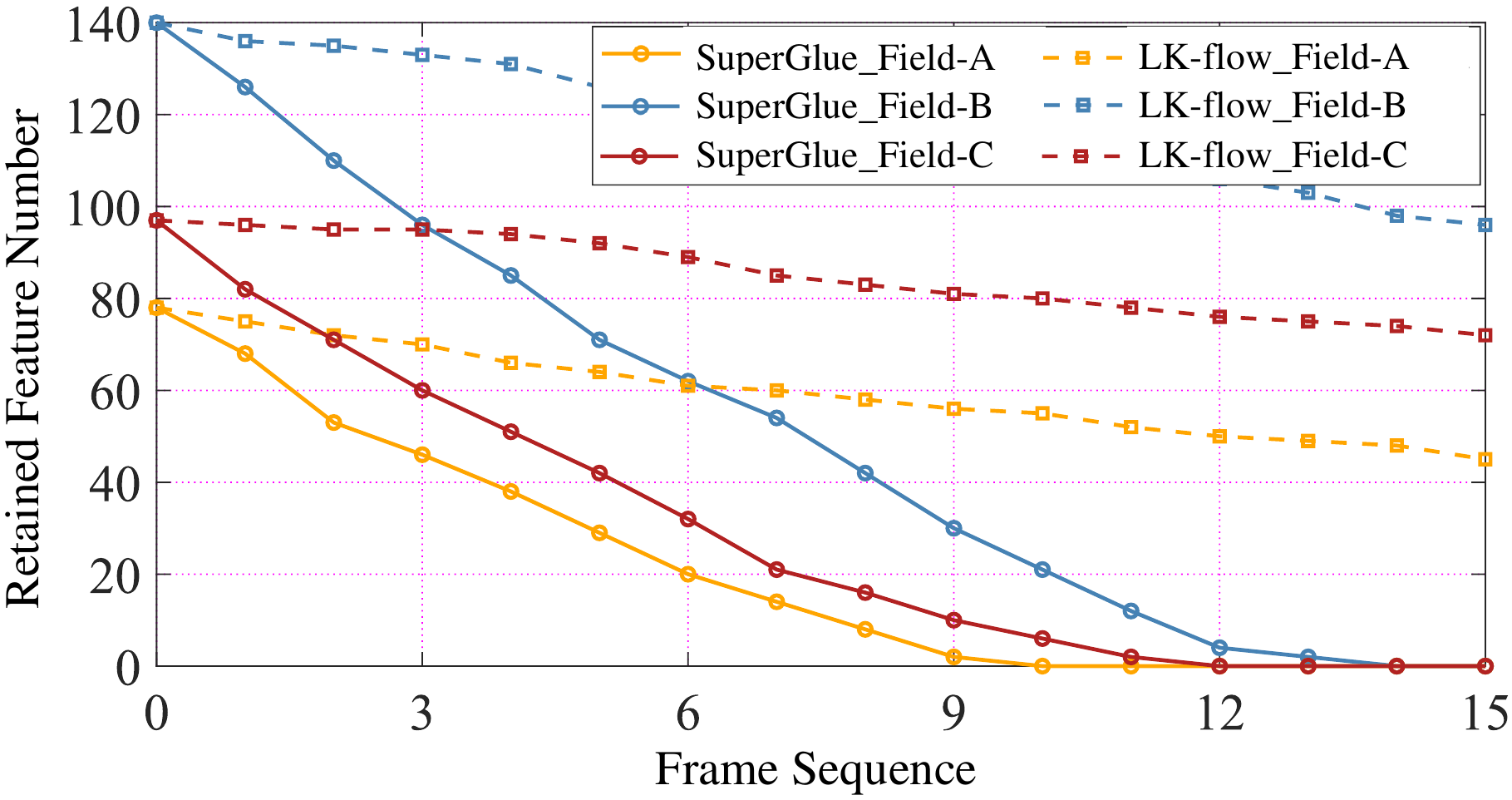}
  \caption{The number of the retained feature along consecutive frame sequence.}
  \label{fig:exp_feature_retained_number}
\end{figure}

To assess feature retention, we compare LK-flow tracking and SuperGlue-based matching over 15 consecutive frames. As shown in Fig.~\ref{fig:exp_feature_retained_number}, features initially detected at frame 0 are tracked using two strategies: (1) continuous LK-flow tracking and (2) repeated SuperPoint detection with SuperGlue matching to the previous frame. We report retention statistics from UAV0 during stable forward flight in Fields A, B, and C scenes.

LK-flow demonstrates strong feature inheritance, retaining over 53\% of the initial features across 15 consecutive frames and significantly increasing the number of observations. In contrast, SuperGlue struggles to preserve features over time, with none retained beyond 14 frames. This limitation is primarily due to SuperPoint's low repeatability across consecutive frames, leading to rapid degradation in feature continuity.

In summary, leveraging the guidance-prediction design, the GP-SS method significantly enhances cross-camera feature association frequency. Furthermore, the integration of LK-flow ensures effective feature inheritance across consecutive frames in each camera, thereby guaranteeing sufficient multi-view observations.

\begin{figure*}[]
  \centering
  \includegraphics[width=1.0\linewidth]{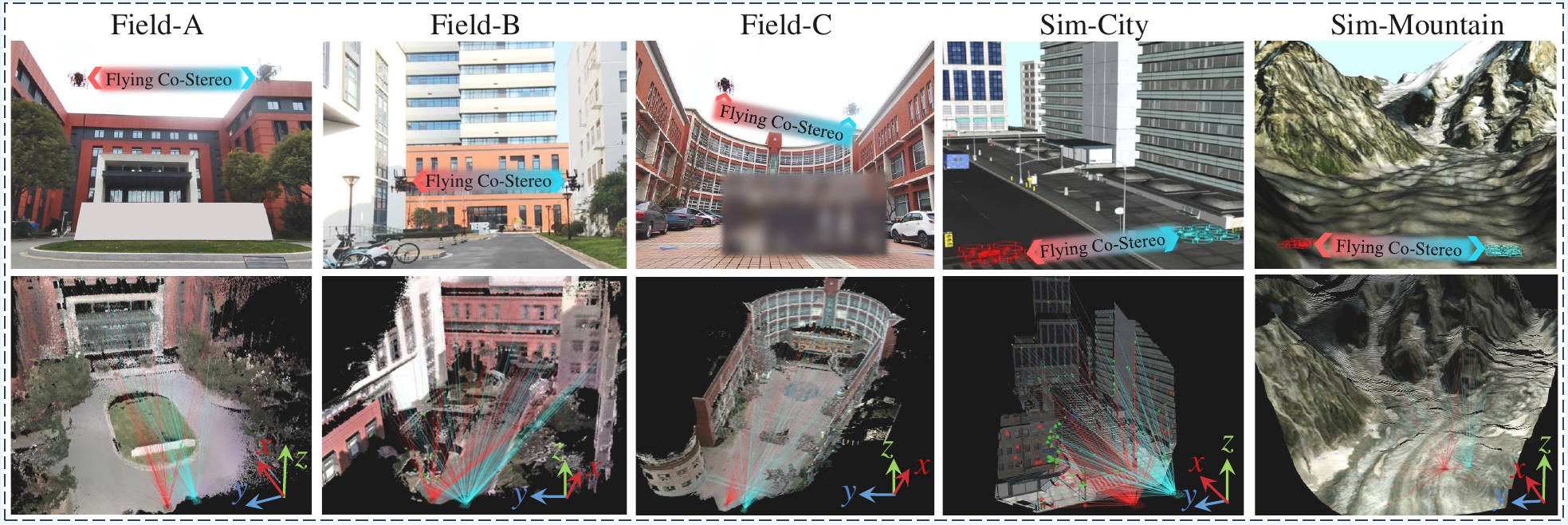}
  \caption{Evaluation scenarios of the Flying Co-Stereo system in both real-world and simulated environments.}
  \label{fig:exp_scene_real_sim}
\end{figure*}

\begin{table*}[]
  \caption{The Experimental Parameters of Real-world and Simulated Large-scale Scenes}
  \label{tab:experiment_parameter}
  \centering
  \setlength{\tabcolsep}{5.2mm} 
  \renewcommand{\arraystretch}{1.2} 
  \begin{tabular}{|c|c|c|c|c|c|}
  \hline
                                Datasets & Field-A                                        & Field-B                                          & Field-C                                          &        Sim-City                                                         & Sim-Mountain                                                    \\ \hline
  Scene Size (L*W*H)(m)          & 45*38*28                                                              &                     68*40*45                                         & 70*30*25                                                            & 43*15*35                                                            & 130*150*60                                                          \\ \hline
  Scene Depth (Avg/Max) (m)       & 35/45    & 36/68    & 40/70    & 30/43    & 60/130   \\ \hline
  Flying Co-Stereo Baseline (m)         & 2.88     & 2.48    & 3.08   & 3.00   & 3.00   \\ \hline
  UAV0 Camera Translation (m)     & \begin{tabular}[c]{@{}c@{}}x:0.412\\ y:-0.024\\ z:0.025\end{tabular} & \begin{tabular}[c]{@{}c@{}}x:0.409\\ y:-0.112\\ z:-0.052\end{tabular} & \begin{tabular}[c]{@{}c@{}}x:0.422\\ y:0.034\\ z:0.015\end{tabular} & \begin{tabular}[c]{@{}c@{}}x:0.400\\ y:0.000\\ z:0.000\end{tabular} & \begin{tabular}[c]{@{}c@{}}x:1.000\\ y:0.300\\ z:0.800\end{tabular} \\ \hline
  UAV1 Camera Translation (m)     & \begin{tabular}[c]{@{}c@{}}x:0.410\\ y:-0.055\\ z:0.032\end{tabular}& \begin{tabular}[c]{@{}c@{}}x:0.403\\ y:-0.095\\ z:-0.042\end{tabular} & \begin{tabular}[c]{@{}c@{}}x:0.408\\ y:0.023\\ z:0.012\end{tabular} & \begin{tabular}[c]{@{}c@{}}x:0.400\\ y:0.000\\ z:0.000\end{tabular} & \begin{tabular}[c]{@{}c@{}}x:1.000\\ y:0.300\\ z:0.800\end{tabular} \\ \hline
  Num of Camera Frames (UAV0 / UAV1)   & 4/4      & 4/4      & 4/4     & 4/4     & 10/10   \\ \hline
  Translation-to-Depth Ratio (\%) & \begin{tabular}[c]{@{}c@{}}UAV0:1.21\\ UAV1:1.22\end{tabular}  &  \begin{tabular}[c]{@{}c@{}}UAV0:1.15 \\ UAV1:1.14\end{tabular} & \begin{tabular}[c]{@{}c@{}}UAV0:1.05\\ UAV1:1.02\end{tabular}       & \begin{tabular}[c]{@{}c@{}}UAV0:1.33\\ UAV1:1.33\end{tabular}       & \begin{tabular}[c]{@{}c@{}}UAV0:2.88\\ UAV1:2.88\end{tabular}       \\ \hline
  Baseline-to-Depth Ratio (\%)    & 8.2      & 6.9      & 7.7     & 10.0     & 5.0   \\ \hline
  Num of Total Landmarks         & 78    & 140    & 97    & 135   & 85   \\ \hline
  \end{tabular}
\end{table*}

\subsection{Collaborative Triangulation of Sparse Landmarks} 
Compared to single-UAV landmark triangulation, the collaborative two-UAV system introduces an additional camera $C_1$ from UAV1. This section systematically analyzes how $C_1$ affects four key aspects of 3D landmark estimation: (1) landmark accuracy and distribution, (2) solution stability, (3) sensitivity to dynamic baseline variations, and (4) the selection of optimal baseline length for accurate triangulation.

\subsubsection{\textbf{Landmark Accuracy and Distribution}} We conduct simulated and real-world experiments to evaluate the accuracy and depth distribution of 3D landmarks. Fig.~\ref{fig:exp_scene_real_sim} illustrates our experimental setups. For real-world ground truth reference, we employ R3LIVE\cite{lin2022R3live} using Lidar for high-fidelity scene reconstruction. The reconstruction error is quantified using the Closest Point Error (CPE), which measures the Euclidean distance between each reconstructed sparse 3D landmark and its nearest neighbor in the ground truth point cloud. Table~\ref{tab:experiment_parameter} summarizes key experimental parameters, including: scene dimensions, Flying Co-Stereo baseline length and movements, as well as the number of camera frames. We also triangulate the landmarks from a single perspective of UAV0 for comparison. Table~\ref{tab:sparse_lmd_in_depths} compares the number of reconstructed landmarks and the average reconstruction error (CPE) between the Flying Co-Stereo system and a single UAV across different depth segments. We divide the depth range into four segments: 0$\sim$10 m, 10$\sim$30 m, 30$\sim$50 m, and 50$\sim$70 m. 

The Flying Co-Stereo system integrates single-UAV near-field VIO landmarks with co-visible long-range landmarks, resulting in more landmarks across all depth ranges compared to UAV0 alone. Notably, UAV0 fails to triangulate landmarks beyond 30 m, whereas Flying Co-Stereo maintains effective triangulation capability at such distances. In terms of reconstruction accuracy, the single-UAV system performs better for near-field landmarks within 10 m, primarily due to baseline pose errors in the Flying Co-Stereo setup, which degrade the triangulation accuracy of nearby co-visible landmarks. However, for depths beyond 10 m, the wide-baseline advantage of Flying Co-Stereo becomes prominent, leading to significantly improved triangulation accuracy over the single-UAV approach.

\begin{table}[]
  \caption{The number of valid landmarks in different depth segments by Flying Co-Stereo and single UAV0}
  \label{tab:sparse_lmd_in_depths}
  \setlength{\tabcolsep}{1.5mm} % 调整列间距
  \begin{tabular}{lccccc}
  \toprule
  \multirow{2}{*}{Datasets} & \multirow{2}{*}{Config} & \multicolumn{4}{c}{Number of Landmark with CPE (m)}      \\ \cline{3-6} 
                            & & 0$\sim$10m & 10$\sim$30m & 30$\sim$50m & 50$\sim$70m \\ \hline
  \multirow{2}{*}{Field-A}  & Co-Stereo & 27 (0.18)  & 5 (0.44)  & 46 (0.78)  & \textcolor{red}{$\times$}                       \\
                            & UAV0 & 13 (0.17)  & \textcolor{red}{$\times$}   & \textcolor{red}{$\times$}   & \textcolor{red}{$\times$}                       \\ \hline
  \multirow{2}{*}{Field-B} & Co-Stereo & 33 (0.28)   & 52 (0.53)   & 27 (0.88)   & 28 (1.12)                       \\
                            & UAV0 & 25 (0.25)   & \textcolor{red}{$\times$}  & \textcolor{red}{$\times$}   & \textcolor{red}{$\times$}                       \\ \hline
  \multirow{2}{*}{Field-C} & Co-Stereo & 14 (0.23)   & 20 (0.46)  & 20 (0.67)  & 43 (0.97)                      \\
                            & UAV0 & 13 (0.21)  & 4 (2.32)  & \textcolor{red}{$\times$}   & \textcolor{red}{$\times$}                       \\ \hline
  \multirow{2}{*}{Sim-City} & Co-Stereo & 50 (0.10)  & 51 (0.35)  & 26 (0.85)  & 8 (1.02)                      \\
                            & UAV0 & 16 (0.11)  & 12 (1.23)  & \textcolor{red}{$\times$}  & \textcolor{red}{$\times$}                       \\ \hline
  \multirow{2}{*}{Sim-Mountain} & Co-Stereo & 21 (0.21)  & 31 (0.43)  & 18 (0.76)  & 15 (1.26)                      \\
                                & UAV0 & 15 (0.19)  & 17 (1.34)  & \textcolor{red}{$\times$}   & \textcolor{red}{$\times$}                       \\ \bottomrule
  \end{tabular}
  \end{table}

\subsubsection{\textbf{Stability Analysis of Triangulation}} 
\begin{figure}[]
  \centering
  \includegraphics[width=1.0\linewidth]{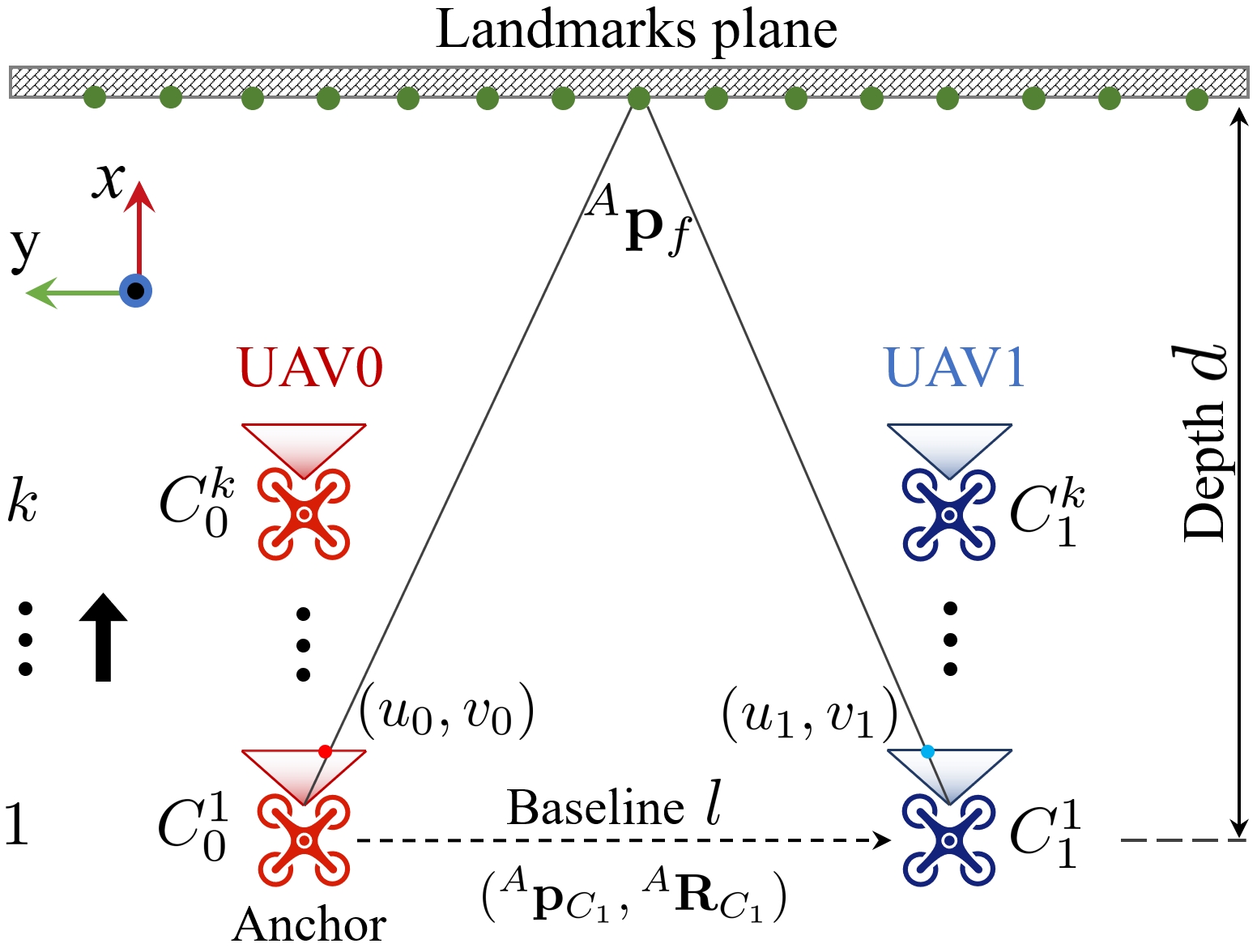}
  \caption{The triangulation configurations for two UAVs flying in parallel.}
  \label{fig:exp_triangulation}
\end{figure}

We further assess triangulation stability through condition number analysis. Considering that typical UAV missions prioritize forward motion toward targets rather than lateral movement, we design experiments where two UAVs fly in parallel toward a destination ahead. This setup allows us to investigate how flight distance and baseline length influence triangulation stability. As shown in Fig.~\ref{fig:exp_triangulation}, two UAVs fly forward across $k$ keyframes with baseline length $l$, while landmarks are uniformly distributed on a 20 m $\times$ 20 m plane at depth $d = 30$ m, with 0.5 m spacing.

\begin{figure}[]
  \centering
  \includegraphics[width=1.0\linewidth]{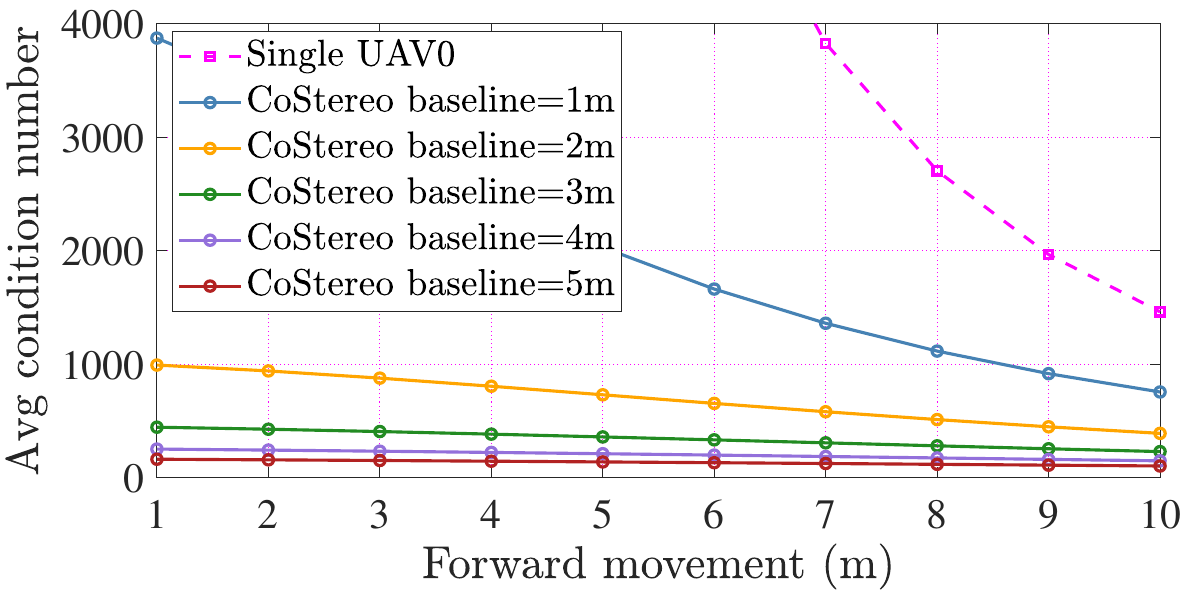}
  \caption{Comparison of the average condition number of landmarks triangulation between the Flying Co-Stereo system (with different baseline lengths $l$) and single UAV0.}
  \label{fig:exp_condition_number}
\end{figure}

In our experiment configuration, the forward movement spans from 1 m to 10 m, with a 0.1 m keyframe step. For the Flying Co-Stereo system, the baseline varies from 1 m to 5 m. Figure~\ref{fig:exp_condition_number} presents an analysis of the average triangulation condition numbers (computed using Eq.\eqref{eq:condition_number}) for all planar landmarks across varying baseline lengths with forward movement distances. The results show that the Flying Co-Stereo system achieves significantly lower condition numbers compared to the single-UAV system with the same forward movement. A lower condition number indicates greater triangulation stability and reduced sensitivity to noise.

Moreover, the analysis across different baseline lengths shows that increasing the baseline from 1 m to 2 m yields substantial improvements in triangulation stability, while further extending the baseline from 2 m to 5 m results in diminishing returns.

Finally, the results indicate that increasing forward movement offers limited improvements in the triangulation stability for the Flying Co-Stereo system. This suggests that the system primarily relies on its wide baseline advantage to achieve stable long-range triangulation, rather than depending heavily on extensive forward motion. Such behavior is consistent with stereo vision principles, contrasting with structure-from-motion (SfM) approaches.

\subsubsection{\textbf{Sensitivity to Baseline Variations}} We further investigate the sensitivity of landmark reconstruction errors with the dynamic baseline errors. To simplify the problem, we reduce the triangulation window length to 1, utilizing only a single pair of $C_0^1$ and $C_1^1$ cameras at the initial timestamp. We remove the time index for simplicity. The spatial configuration of the two UAVs and landmarks is illustrated in Fig.~\ref{fig:exp_triangulation}. The pixel observations of $^A\mathbf{p}_f$ in cameras $C_0$ and $C_1$ are denoted as ($u_0,v_0$) and ($u_1,v_1$), respectively. The corresponding observation bearing vectors are ${}^A \mathbf{b}_{C_0} = (u_0,v_0,1)^{\top}$ and ${}^A \mathbf{b}_{C_1} = {}^A \mathbf{R}_{C_1} (u_1,v_1,1)^{\top}$. The orthogonal space of bearing vector can be expressed as ${}^A \mathbf{N}_0 = \left\lfloor{ }^A \mathbf{b}_{C_0} \times\right\rfloor$ and ${}^A \mathbf{N}_1 = \left\lfloor{ }^A \mathbf{b}_{C_1} \times\right\rfloor$, respectively.
The stacked measurement of Eq.\eqref{eq:triangulation} is reformulated as follows:

\begin{equation}
  \underbrace{\left[\begin{array}{c}
    {}^A \mathbf{N}_0 \\
    {}^A \mathbf{N}_1 \\
    \end{array}\right]}_{\mathbf{A}_{6\times3}} {}^A \mathbf{p}_f = \underbrace{\left[\begin{array}{c}
    {}^A \mathbf{N}_0 {}^A \mathbf{p}_{C_0} \\
    {}^A \mathbf{N}_1 {}^A \mathbf{p}_{C_1} \\
    \end{array}\right]}_{\mathbf{b}_{6\times1}}
\end{equation}

\begin{figure}[]
  \centering
  \includegraphics[width=1.0\linewidth]{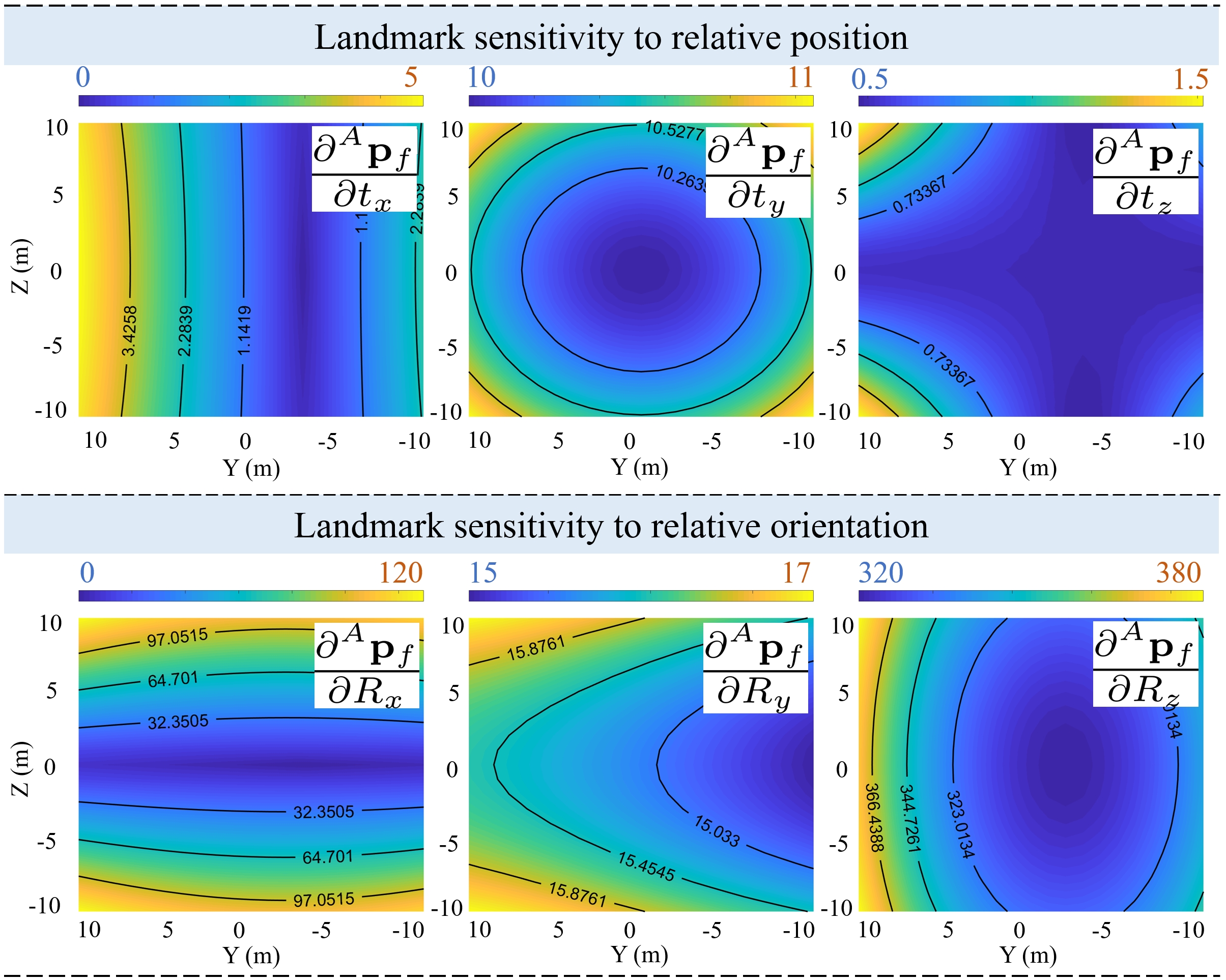}
  \caption{The landmark triangulation sensitivity with respect to relative position and orientation.}
  \label{fig:exp_sensitivity}
\end{figure}

If we define $C_0$ as the anchor camera. ${}^A \mathbf{p}_{C_0}$, which represents the position of $C_0$ in the anchor camera frame, becomes zero. The solution of ${}^A \mathbf{p}_f$ could be expressed as:
\begin{equation}
  {}^A \mathbf{p}_f = \frac{{{}^A \mathbf{N}_1}^{\top} {}^A \mathbf{N}_1 {}^A \mathbf{p}_{C_1}}{{{}^A \mathbf{N}_0}^{\top} {}^A \mathbf{N}_0 + {{}^A \mathbf{N}_1}^{\top} {}^A \mathbf{N}_1}
  \label{eq:triangulation_simple}
\end{equation}

The sensitivity of landmark position ${}^A\mathbf{p}_f$ to baseline variations is quantified by gradient analysis of baseline error perturbations. Let relative position ${}^A \mathbf{p}_{C_1}$ and orientation ${}^A \mathbf{R}_{C_1}$ (derived from ${{}^A \mathbf{N}_1} = \left\lfloor {}^A \mathbf{R}_{C_1} (u_1,v_1,1)^{\top} \times\right\rfloor$) be parameterized by $[t_x, t_y, t_z, R_x, R_y, R_z]^{\top}$. The gradient vector $[\frac{\partial ^A\mathbf{p}_f}{\partial t_x}, \frac{\partial ^A\mathbf{p}_f}{\partial t_y}, \frac{\partial ^A\mathbf{p}_f}{\partial t_z}, \frac{\partial ^A\mathbf{p}_f}{\partial R_x}, \frac{\partial ^A\mathbf{p}_f}{\partial R_y}, \frac{\partial ^A\mathbf{p}_f}{\partial R_z}]^{\top}$ is computed via numerical perturbation. For clarity in visualization, we establish the anchor camera $C_0$ pose at origin $(0,0,0)$ with an identity orientation matrix. The $C_1$ is positioned parallel at $(0,-3,0)$, maintaining a baseline ${}^A \mathbf{p}_{C_1} = (0,-3,0)$ and identity relative orientation ${}^A \mathbf{R}_{C_1} = \mathbf{I}$.

Figure~\ref{fig:exp_sensitivity} presents the sensitivity of the reconstruction error of landmarks on the 20 m $\times$ 20 m landmark plane with a depth of 30 m with respect to both relative positional and orientational errors. Overall, relative orientation errors demonstrate a substantial impact on landmark accuracy compared to relative position errors. In particular, yaw estimation errors exhibit the most pronounced effect (exceeding 320) on landmark reconstruction accuracy. Our Bidirectional View Differential (BVD) algorithm is specifically designed to enhance yaw estimation accuracy. Regarding relative position errors, the sensitivity along the baseline direction substantially exceeds that in other directions. Furthermore, the internal sensitivity distribution pattern in the landmark plane reveals that, among landmarks ahead, those positioned centrally between the two cameras exhibit lower sensitivity to baseline error, while those farther from the center show increased sensitivity.

\subsubsection{\textbf{Optimal Baseline Length for Accurate Triangulation}} We investigate how to select an optimal baseline length to minimize the overall triangulation error of landmarks. A longer baseline is validated to reduce the condition number of triangulation, thereby improving both stability and precision. However, increasing the baseline also enlarges the inter-UAV visual observation distance, which particularly degrades the accuracy of visual PnP-based position estimation, leading to larger errors in baseline position estimation. 

\begin{figure}[]
  \centering
  \includegraphics[width=1.0\linewidth]{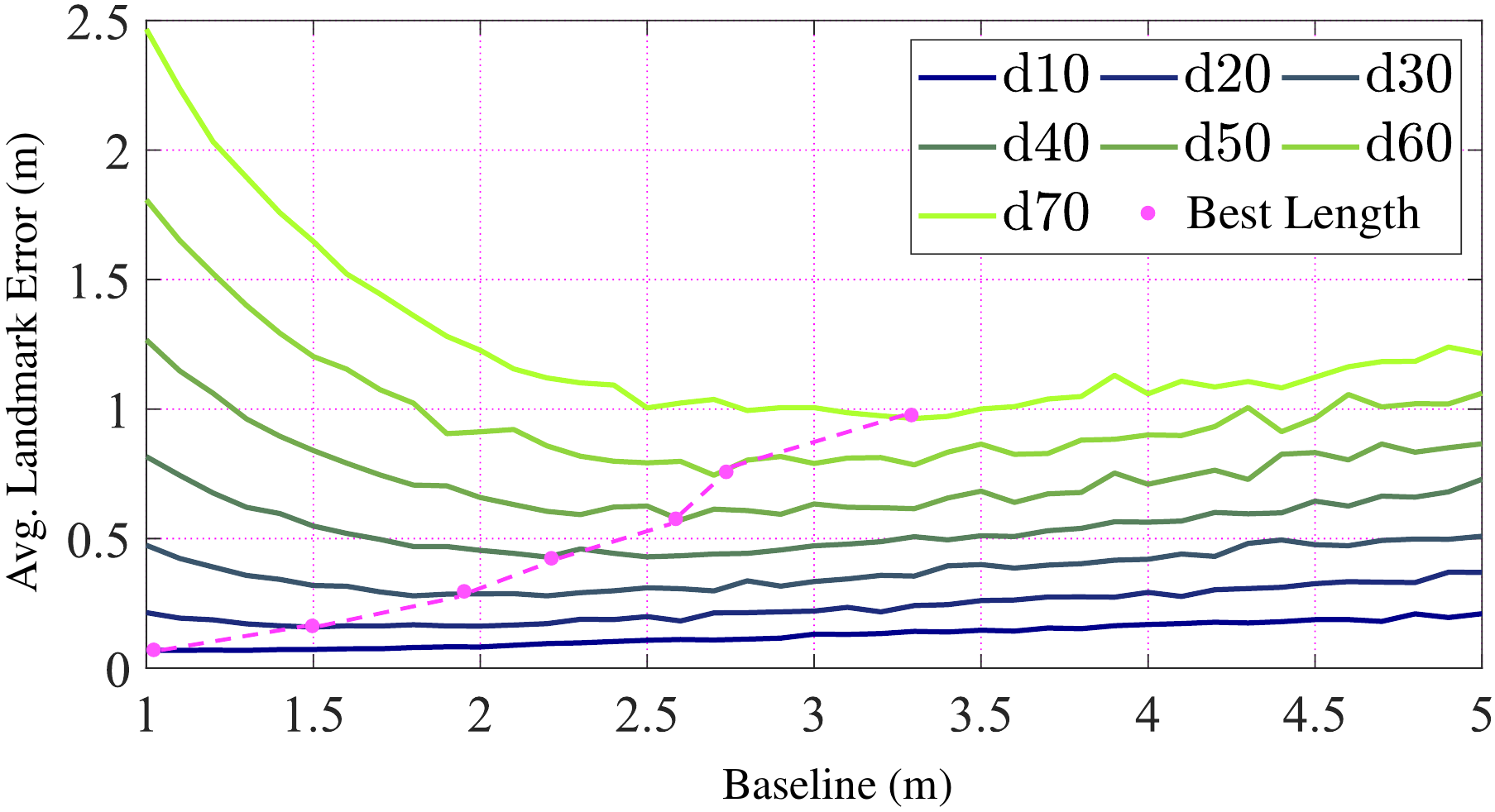}
  \caption{The numerical analysis of landmark triangulation error with respect to baseline length and scene depth. The focal length of the front and side cameras is selected as 380.}
  \label{fig:exp_baseline_search}
\end{figure}

We first model the baseline-dependent PnP position noise, and then assess the effect of baseline position error on the reconstruction errors of landmarks using the triangulation Eq.\eqref{eq:triangulation_simple}.
For the PnP noise model, Gaussian pixel noises $\Delta u \sim \mathcal{N}(0, 1), \Delta v \sim \mathcal{N}(0, 1)$ are added to the 2D pixel observations of the IR markers on UAV1, captured by the side camera on UAV0. Based on the perspective camera model, the induced baseline position error is given by:
\begin{equation}
\Delta {}^{A} \mathbf{p}_{C_1} = (l\frac{\Delta u}{f}, l^2\frac{\sqrt{\Delta u^2 + \Delta v^2}}{f}, l\frac{\Delta v}{f}) 
\label{eq:baseline_noise}
\end{equation}
where $l$ is the baseline length and $f$ is the focal length of the side camera. Landmarks are then triangulated under perturbed baseline configurations, incorporating the simulated baseline noise. The triangulation error is defined as the Euclidean distance between the estimated and ground-truth (noise-free) landmark positions. We evaluate the average error for landmarks uniformly sampled on depth planes, averaged over 100 Monte Carlo trials. Fig.~\ref{fig:exp_baseline_search} presents the results across baseline lengths from 1 m to 5 m and landmark depths from 10 m to 70 m.

The results reveal depth-dependent error characteristics. At close ranges ($\leq$ 20 m), shorter baselines offer more accurate baseline estimation, leading to lower reconstruction errors. As depth increases (30-70 m), the optimal baseline length (marked as Best Length) also increases, indicating that moderate baseline extension improves triangulation robustness. However, overly long baselines introduce larger estimation errors, which degrade triangulation accuracy.

This reflects a trade-off between stereo geometry parallax and baseline accuracy. In practice, baseline length should be selected based on pixel noise and scene depth distribution. In our experiments, baselines between 2 m and 4 m strike a good balance between parallax and estimation reliability.

\subsection{Long-Range Dense Mapping}
\begin{figure*}[]
  \centering
  \includegraphics[width=1.0\linewidth]{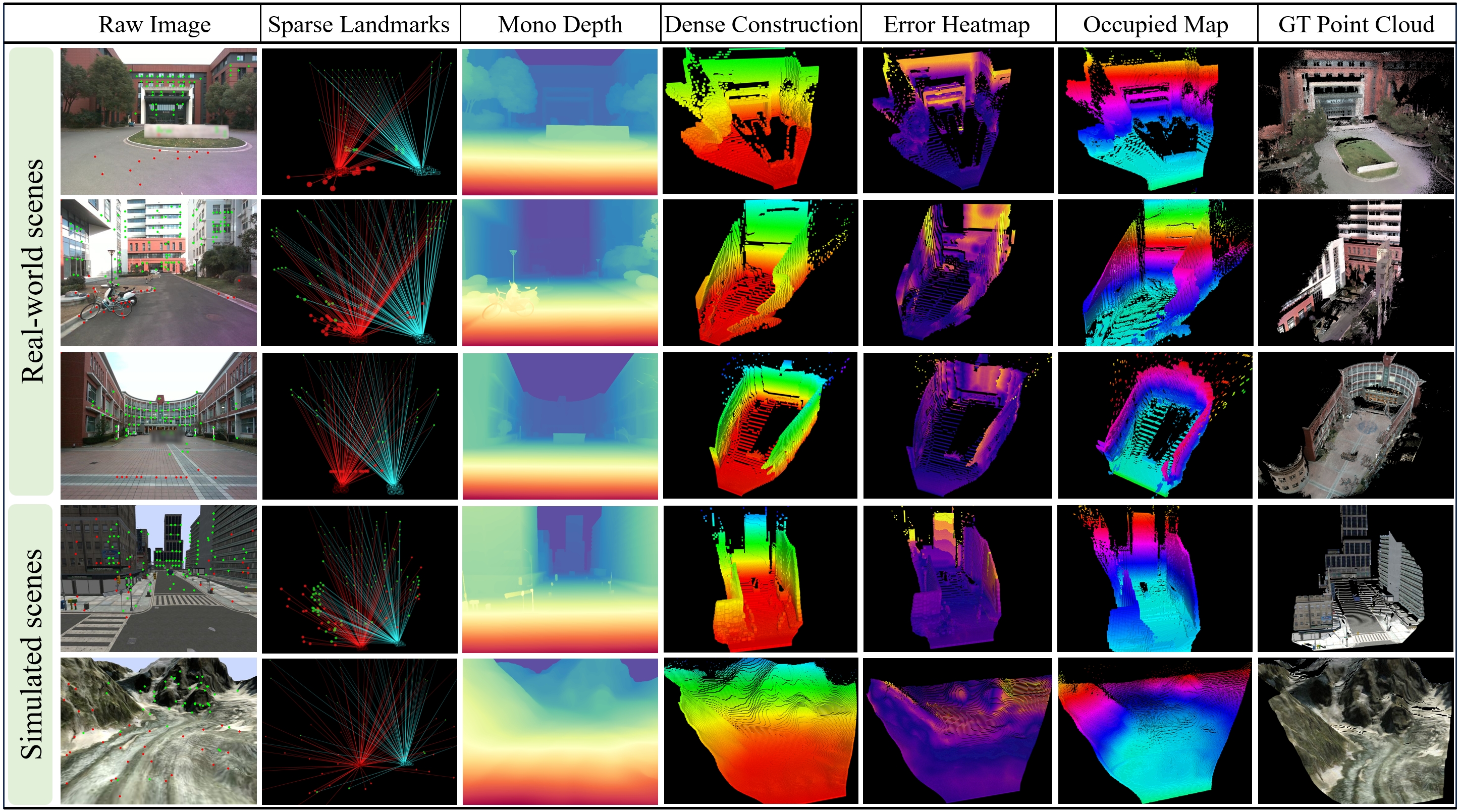}
  \caption{The long-range dense reconstruction experiments in outdoor environments and photorealistic simulation. The visualization comprises seven columns: (1) UAV0's current image with VIO features (red) and co-visible features (green); (2) Sparse landmarks from collaborative triangulation and self VIO process; (3) Metric-ambiguous depth from DepthAnythingV2\cite{yang2024DepthAnythingV2}; (4) Dense reconstruction via exponential fitting of sparse landmarks and depth map, resampled into a dense point cloud; (5) Heatmap of reconstruction error; (6) Occupancy grid from Voxblox\cite{oleynikova2017voxblox} fusion of multi-frame point clouds; and (7) Ground-truth point clouds from R3LIVE\cite{lin2022R3live} (real-world) or Gazebo (simulated).}
  \label{fig:exp_dense_mapping}
\end{figure*}

This section first presents the long-range dense mapping results across multiple scenes, followed by an analysis of reconstruction accuracy. Subsequently, we compare the performance of different fitting models for large-scale reconstruction and investigate how sparse feature distribution influences dense reconstruction quality. Finally, two advanced MVS methods are included as benchmarks for comparison. To comprehensively evaluate the system, we conduct various real-world experiments alongside simulated scenarios in urban streets and mountainous environments to enrich diversity. Relative baseline noise is injected according to Eq.\eqref{eq:baseline_noise} in the simulation. Both the real-world and simulated experimental results are illustrated in Fig.~\ref{fig:exp_dense_mapping}.

\subsubsection{\textbf{Accuracy of Dense Reconstruction}}
Since the ground truth point cloud from R3LIVE\cite{lin2022R3live} contains extraneous sampled regions (outside our evaluation areas), we compute the unidirectional Chamfer distance (uCD) from the predicted point cloud to the ground truth to focus only on relevant regions. The uCD metric is defined as:
\begin{equation}
  uCD_{(P \to G)} = \frac{1}{|P|} \sum_{p \in P} \min_{g \in G} \| p - g \|_2
\end{equation}
where $P$ and $G$ represent the predicted and ground truth point clouds, respectively.
As shown in Table~\ref{tab:uCD_dense_mapping}, the unidirectional Chamfer distance (uCD) is computed both overall and across four depth intervals: 0$\sim$10 m, 10$\sim$30 m, 30$\sim$50 m, and 50$\sim$70 m. The experimental results show that reconstruction errors maintain $<$0.5 m accuracy within a depth of 10 m. The error progressively escalates as depth increases. The relative error is computed as the ratio of the uCD to the average depth of the corresponding depth segment, which ranges from 2.3\% to 9.7\% across four depth segments. The results demonstrate that the proposed Flying Co-Stereo system is capable of achieving accurate dense reconstruction in large-scale, long-range environments.

\begin{figure}[]
  \centering
  \includegraphics[width=1.0\linewidth]{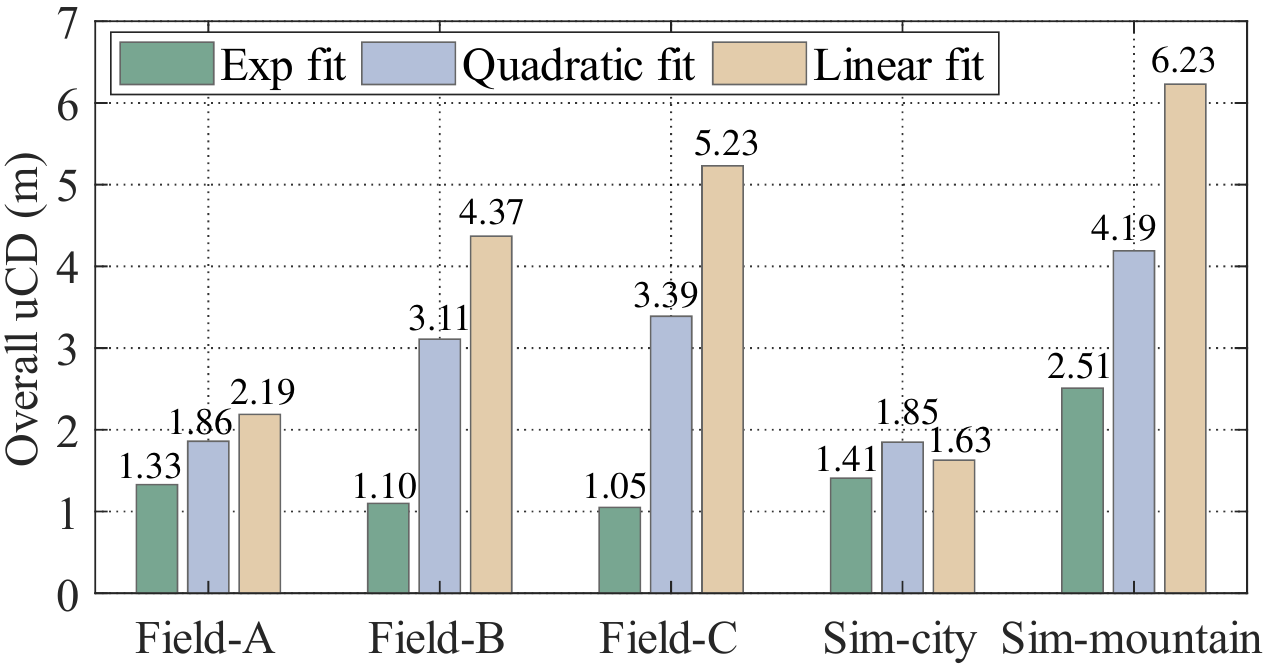}
  \caption{Overall uCD comparison of different fitting methods across multiple scenes.}
  \label{fig:exp_error_with_diff_fit}
\end{figure}

\begin{table}[]
  \caption{The uCD of Predicted Dense Point Cloud with Ground Truth}
  \label{tab:uCD_dense_mapping}
  \setlength{\tabcolsep}{1.8mm} % 调整列间距
  \begin{tabular}{lcccc}
  \toprule
  \multirow{2}{*}{Datasets} & \multicolumn{4}{c}{uCD in Depth Segment (m) with relative error (\%)}              \\ \cline{2-5} 
                            & 0$\sim$10m & 10$\sim$30m & 30$\sim$50m  & 50$\sim$70m \\ \hline
  Field-A                   & 0.22 \textcolor{deepgreen}{(4.4\%)}   & 1.12 \textcolor{deepgreen}{(6.1\%)}   & 1.65 \textcolor{deepgreen}{(4.1\%)}   & -                           \\
  Field-B                 & 0.48 \textcolor{deepgreen}{(9.6\%)}   & 1.23 \textcolor{deepgreen}{(6.2\%)}    & 2.91 \textcolor{deepgreen}{(7.2\%)}   & 1.89 \textcolor{deepgreen}{(3.2\%)}                      \\
  Field-C                  & 0.34 \textcolor{deepgreen}{(6.8\%)}   & 0.74 \textcolor{deepgreen}{(3.7\%)}   & 0.95 \textcolor{deepgreen}{(2.3\%)}  & 1.82 \textcolor{deepgreen}{(3.0\%)}                       \\
  Sim-City                  & 0.46 \textcolor{deepgreen}{(9.2\%)}   & 1.62 \textcolor{deepgreen}{(8.1\%)}   & 2.61 \textcolor{deepgreen}{(6.5\%)}   & 4.83 \textcolor{deepgreen}{(8.1\%)}                       \\
  Sim-Mountain              & 0.43 \textcolor{deepgreen}{(8.6\%)}      & 1.95 \textcolor{deepgreen}{(9.7\%)}       & 2.46 \textcolor{deepgreen}{(6.2\%)}      & 2.71 \textcolor{deepgreen}{(4.5\%)}                          \\ \bottomrule
  \end{tabular}
  \end{table}

\begin{figure*}[]
  \centering
  \includegraphics[width=1.0\linewidth]{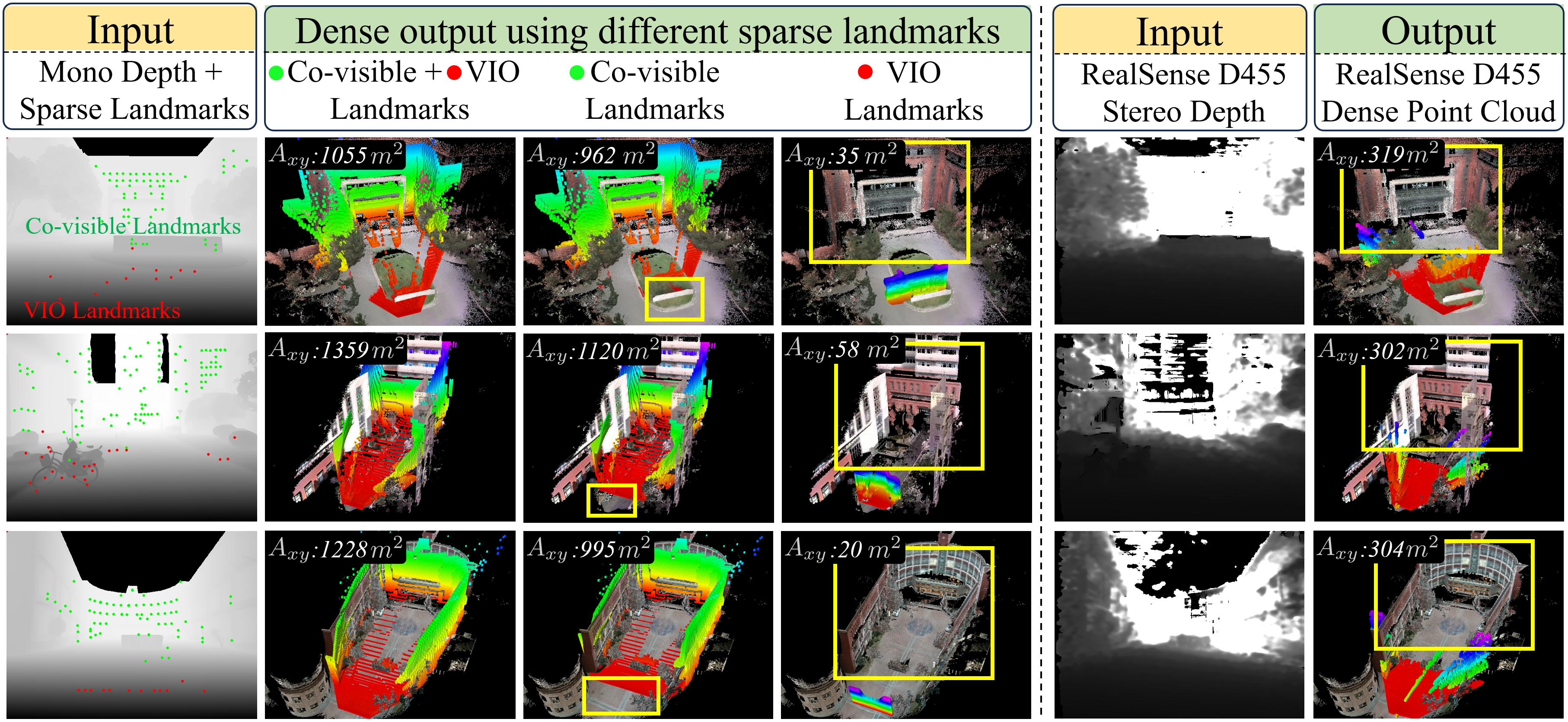}
  \caption{Comparison of dense reconstruction results from the sparse-to-dense pipeline of Flying Co-Stereo and a commercial stereo camera. The dense coverage area $A_{xy}$ is further evaluated using different types of sparse landmarks. Yellow boxes highlight regions with missing reconstruction coverage.}
  \label{fig:exp_compare_stereo_dense}
\end{figure*}

\subsubsection{\textbf{Comparison of Fitting Methods}} We also compare the reconstruction accuracy based on different fitting models, including linear, quadratic, and exponential fitting. As shown in Fig.~\ref{fig:exp_error_with_diff_fit}, the exponential fitting model consistently outperforms the others across multiple scenes, achieving uCD values of 1.33 m (Field-A), 1.10 m (Field-B), 1.05 m (Field-C), 1.41 m (Sim-City), and 2.51 m (Sim-Mountain) - representing average 67\% and 55\% error reduction compared to linear and quadratic approaches. These results highlight the exponential model’s effectiveness in large-scale, long-range dense reconstruction.

\subsubsection{\textbf{Landmark Distribution Impact}} Our analysis further investigates how the distribution of sparse landmarks, combined with exponential fitting, influences dense reconstruction coverage. As illustrated in Fig.~\ref{fig:exp_compare_stereo_dense}, we categorize sparse landmarks into three groups: (1) Co-visible + VIO landmarks, (2) Co-visible-only landmarks, and (3) VIO-only landmarks. To quantify spatial coverage, the reconstructed point clouds are first projected onto the XY plane. The 2D coverage area $A_{xy}$ is then estimated as the area of the convex hull formed by the projected point clouds, providing an approximation of the system's effective sensing range. The computed coverage areas are labeled in Fig.~\ref{fig:exp_compare_stereo_dense}.

Comparative analysis reveals distinct characteristics across the groups. The Co-visible + VIO group achieves the most comprehensive coverage across all depth ranges, with coverage areas of 1055 $m^2$ (Field-A), 1359 $m^2$ (Field-B), and 1228 $m^2$ (Field-C). The Co-visible-only group exhibits coverage gaps in the near field, while the VIO-only group demonstrates limited overall coverage and fails entirely at larger depths. These findings validate the complementary roles of co-visible landmarks (enabling long-range reconstruction) and VIO landmarks (ensuring near-field reconstruction).

\subsubsection{\textbf{Comparison with Conventional Camera}} We further benchmark the system against the RealSense D455 stereo camera (95 mm baseline of two infrared lenses). As shown in Fig.~\ref{fig:exp_compare_stereo_dense}, the D455 exhibits a limited effective range of $\leq$ 20 m with coverage areas of 319 $m^2$ (Field-A), 302 $m^2$ (Field-B), and 304 $m^2$ (Field-C). Areas lacking coverage are marked with yellow boxes. The Flying Co-Stereo achieves 331-450\% greater coverage across test scenarios (Field-A: 331\%, Field-B: 450\%, Field-C: 403\%), demonstrating its superior capability for large-scale perception.

\begin{figure}[]
  \centering
  \includegraphics[width=1.0\linewidth]{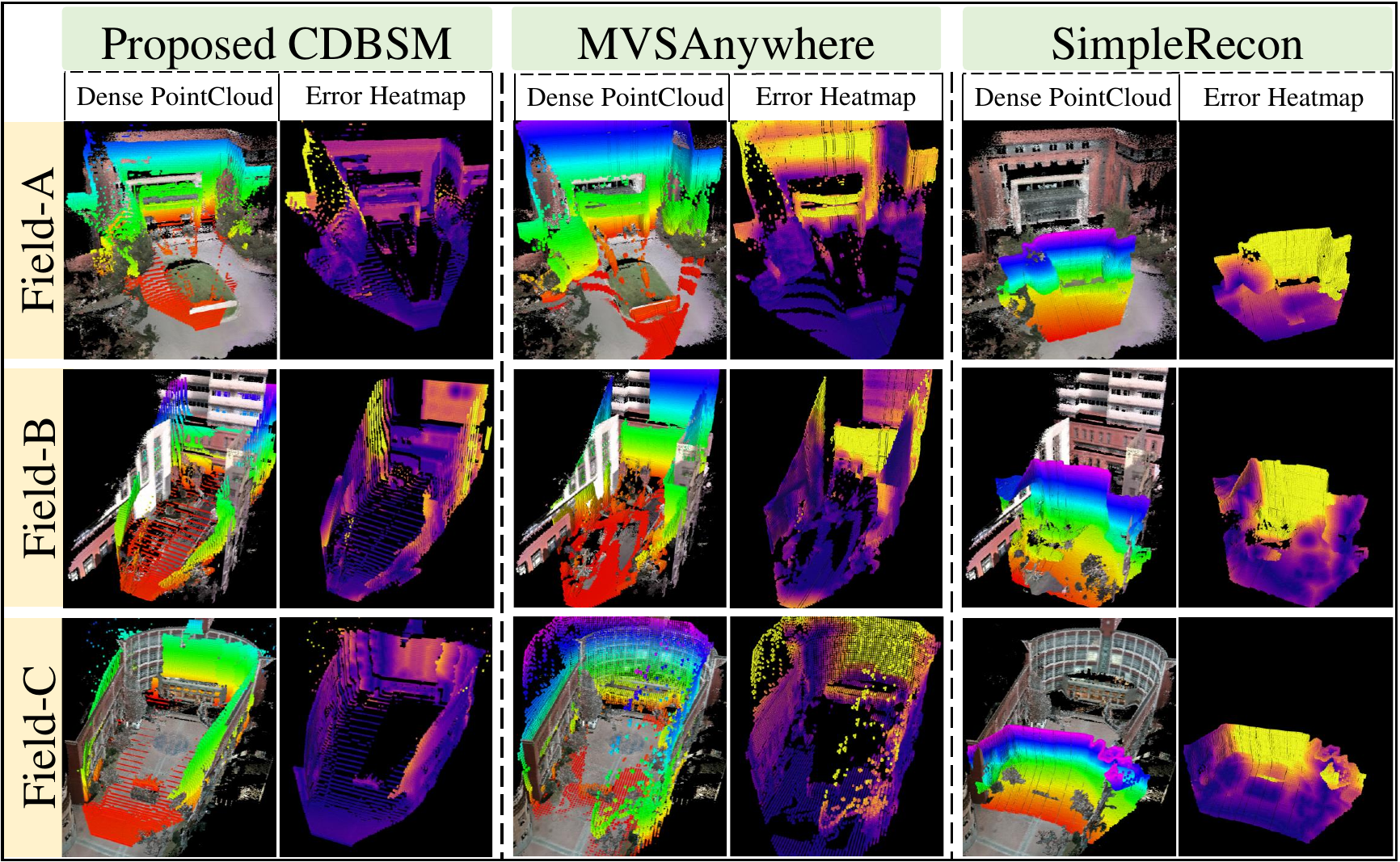}
  \caption{Comparison of reconstruction from MVSAnywhere\cite{izquierdo2025mvsanywhere}, SimpleRecon\cite{sayed2022simplerecon}, with our proposed CDBSM, shown alongside ground truth.}
  \label{fig:exp_compare_MVS}
\end{figure}

\begin{table}[]
  \centering
  \caption{Overall uCD (m) for CDBSM and MVS Methods}
  \label{tab:uCD_MVS}
  \setlength{\tabcolsep}{3.0mm}{
  \begin{tabular}{cccc}
  \toprule
  Datasets    & CDBSM      & MVSAnywhere\cite{izquierdo2025mvsanywhere}   & SimpleRecon\cite{sayed2022simplerecon}  \\ \hline
  Field-A    & \textbf{1.33} m    & 2.38 m    & 13.61 m     \\
  Field-B    & \textbf{1.10} m    & 2.15 m    & 12.32 m      \\
  Field-C    & \textbf{1.05} m    & 2.06 m    & 15.62 m      \\
  \bottomrule
  \end{tabular}
  }
\end{table}

\subsubsection{\textbf{Comparison with MVS Methods}} Figure~\ref{fig:exp_compare_MVS} compares the proposed CDBSM method against two advanced MVS approaches. We compared to SimpleRecon\cite{sayed2022simplerecon}, an accurate MVS method with efficient 2D cost volumes, and MVSAnywhere\cite{izquierdo2025mvsanywhere}, a zero-shot method known for its robustness across any scenes and depth ranges. Both MVS methods utilize two-view images from UAV0 and UAV1 as the reference and source frames, respectively.

While MVSAnywhere performs well in recovering large-scale structures, it tends to lose fine-grained details at long distances. SimpleRecon, on the other hand, exhibits poor scale recovery in long-range scenarios, possibly due to its training on small-scale indoor datasets.
Quantitative results are summarized in Table~\ref{tab:uCD_MVS}. Our proposed CDBSM method consistently outperforms the others, achieving uCD of 1.33 m (Field-A), 1.10 m (Field-B), and 1.05 m (Field-C), representing 44\%(Field-A), 48\%(Field-B) and 49\%(Field-C) error reduction compared to the second-best method, MVSAnywhere.

\section{Conclusion}
This paper introduces Flying Co-Stereo, a collaborative stereo vision system that establishes dynamic wide-baseline configurations using two UAVs for large-scale, long-range dense mapping. Real-world experiments demonstrate that the proposed CDBSM framework enables accurate and robust baseline estimation in complex environments while maintaining efficient feature matching with resource-constrained computers under changing viewpoints, ultimately achieving accurate 70 m-range dense mapping with 2.3-9.7\% relative error. This long-range mapping capability significantly surpasses conventional stereo cameras, achieving up to a 350\% increase in perception range and a 450\% boost in coverage area.

The Flying Co-Stereo system demonstrates promising performance in collaborative aerial perception, offering a practical solution for long-range awareness mapping and navigation tasks. While Flying Co-Stereo demonstrates considerable potential, several directions merit further investigation: (1) Baseline Adaptation: The current system employs a predefined, minimally varying baseline determined by the average scene depth, which may be suboptimal over the entire flight trajectory. Future work will explore active baseline planning strategies that continuously adapt the baseline based on the environmental depth distribution.
(2) Mapping Enhancement: Advancing monocular depth consistency across video frames will further stabilize our sparse-to-dense mapping.
(3) Scalable Architecture: Extension to multi-UAV networks via a serial cascading architecture, where each UAV simultaneously tracks its predecessor and guides its successor, would create an extensible collaborative mapping framework.

\enlargethispage{-\baselineskip} % Reduce the space by one line

% \printbibliography[heading=bibnumbered]
\bibliographystyle{IEEEtran}
\bibliography{ref_dense-depth,ref_multi-robot-collaboration,ref_vio_slam,ref_wide-baseline-system,ref_relative-pose-estimation,ref_image_match,ref_our_lab_paper,ref_sensor-fusion}
% \bibliographystyle{IEEEtran} 
% \begingroup
% \footnotesize
% \printbibliography
% \endgroup

\end{document}